\documentclass[preprint,12pt]{elsarticle}
\usepackage[labelfont=bf,textfont={bf}]{caption}
\usepackage{textcomp}
\usepackage{dblfloatfix}  

\usepackage{mathptmx}
\usepackage{amsmath}
\usepackage{lineno}
\usepackage{balance}
\usepackage[hidelinks]{hyperref} 
\usepackage{picture}
\usepackage{relsize}
\usepackage{multicol}
\usepackage{soul}
\usepackage{enumitem}
\usepackage{algpseudocode}
\setitemize{noitemsep,topsep=0pt,parsep=0pt,partopsep=0pt,leftmargin=*}

\usepackage{graphicx}
\usepackage{caption}
\usepackage{subcaption}
\usepackage{soul}
\usepackage{pgfplots}
\usepackage{epstopdf}
\usepackage{multirow}
\usepackage{listings}
\usepackage{fancybox}
\usepackage{color}
\usepackage{colortbl}
\usepackage{tcolorbox}% http://ctan.org/pkg/tcolorbox
\makeatletter
\newcommand{\mybox}[1]{%
  \setbox0=\hbox{#1}%
  \setlength{\@tempdima}{\dimexpr\wd0+13pt}%
  \begin{tcolorbox}[boxrule=0.5pt, colback=white, arc=4pt,
	        left=6pt,right=6pt,top=6pt,bottom=6pt,boxsep=0pt]
			    #1
  \end{tcolorbox}
}

\definecolor{mygray}{gray}{.9}

\definecolor{mycolorhigh}{RGB}{168, 157, 89}
\definecolor{mycolormiddle}{RGB}{224, 224, 182}
\definecolor{mycolorlow}{RGB}{245, 245, 237}

\newcommand{\IT}{{{FairMatch}}}
\newcommand{\bi}{\begin{itemize}}
\newcommand{\ei}{\end{itemize}}
\renewcommand{\footnotesize}{\scriptsize}
\usepackage{booktabs}
\usepackage[ruled,vlined]{algorithm2e}
\newtcolorbox{blockquote}{colback=blue!5,boxrule=0.4pt,colframe=black,fonttitle=\bfseries}
\begin{document}

% \title{ Ensemble Methods for Better Multi-Objective  Fairness: Experiments with {\IT} }
\title{ Whence Is A Model Fair? Fixing Fairness Bugs \\via Propensity Score Matching}
% \title{Fairer Evaluation Schema for Software Fairness Testing (via Causality Matching)}

\author[1]{Kewen Peng\corref{cor1}}
\ead{kpeng@ncsu.edu}

\author[2]{Yicheng Yang}
\ead{yy546@cornell.edu}

\author[2]{Hao Zhuo}
\ead{hz324@cornell.edu}

\author[1]{Tim Menzies}
\ead{timm@ieee.org}

\address[1]{ North Carolina State University, Raleigh, North Carolina, USA}
% \address[]{Cornell University, Ithaca, New York, USA}
\address[2]{Cornell University, Ithaca, New York, USA}
% \address[]{Department of Computer Science, North Carolina State University, Raleigh, North Carolina, USA}

\cortext[cor1]{Corresponding author}
% \cortext[cor1]{Corresponding author: kpeng@ncsu.edu, yy546@cornell.edu, hz324@cornell.edu, timm@ieee.org,}

% The paper headers
% \markboth{IEEE Transactions on Software Engineering}%
% {Peng \MakeLowercase{\textit{et al.}}: FairMatch+XGBoost for IEEE Journals}

\begin{abstract}

\noindent
\textbf{Context}:
Fairness-aware learning aims to mitigate discrimination against specific protected social groups (e.g., groups categorized based on gender, ethnicity, age, etc.) with minimum cost at predictive performance. Software engineers should study machine learning algorithms since prior studies have reported the pervasive existence of unfair models (as measured by various fairness metrics).

\noindent
\textbf{Objective}: 
We are concerned that different sampling strategies for training/testing data may skew the assessment of the fairness level of a model.
% Consider the standard scheme where a model is fitted to some training set and then evaluated on some testing set.
Since train/test data is often sampled at random, some bias in the training data might still exist in the test data.  Hence, this paper aims to answer the following question:  
{\em How can we know if the reported fairness metrics truly reflect the fairness level of a model?}

\noindent
\textbf{Method}: We propose FairMatch, a post-processing method that uses {\em propensity score matching} to apply different decision thresholds on specific subgroups. We will find, among the test set, matchable control/treatment pairs (indicated by similar propensity scores) with regard to protected attributes. We perform probabilistic calibration for the rest of the testing samples, aiming to tune the
model on different fairness-aware loss functions.

\noindent
\textbf{Results}: The experiments in this paper show that, with propensity score match, we can (a) precisely locate the subset of test data where the prediction model performs unbiased and (b) significantly mitigate bias on the rest of the test data. 

\noindent
\textbf{Conclusion}: 
In conclusion, the use of propensity can help not only conduct fairness testing more comprehensively but also mitigate bias from a model without harming predictive performance.

\end{abstract}


\maketitle
% \IEEEpeerreviewmaketitle
% \IEEEdisplaynontitleabstractindextext

\section{Introduction}\label{intro}
Software embedded with AI/ML has become omnipresent in everyday life, and more comprehensive testing of its quality is in desperate need. As machine learning software has been introduced to various sensitive domains where the subject of the model is humans (e.g., healthcare, education, employment), it is important to ensure that the decision-making models are under no exposure to any discrimination, intentional or unintentional. 

 Various fairness measurements are defined, and guidelines for fairness testing are promoted. While the fairness metrics are motivated and defined intuitively, we are concerned that the guidelines
 for using these metrics are incomplete.   Consider the
standard scheme where a model is fitted to some training set and then evaluated on some testing set. This train/test data is often selected at random, which
means that (a)~some bias in the training data might be missing the test data; (b)~the reported
levels of fairness from such a study can be inaccurate.
 
 % Traditional testing schema for performance metrics (e.g., accuracy, precision, recall) are rather simple. In many cases, multi-fold cross-validation would suffice. However, such testing schema may become arguable when it comes to fairness testing. One of the major assumptions in fairness-aware learning is that training data contain unintended bias/discrimination for various causes. As a matter of fact, many bias mitigation methods are based on designs that transform the training data before fitting a model. In this spirit, it appears counter-intuitive when one uses the testing data off-the-shelf to compute the fairness metrics: if the training set is assumed to contain bias, the randomly split test set is as likely as the training set to contain bias, thus can not represent the expected distribution of future data. 

This paper attempts to improve these fairness testing guidelines by proposing a causality-driven testing method for evaluating software fairness. We assume:
\begin{itemize}
    \item Biased or imbalanced training data is one of the major reasons for unfairness in machine learning models.
    \item Since traditional train-test-split presumes that the training and testing set share similar data distribution, the testing set also shares a likelihood of bias similar to the training set.
\end{itemize}
The first presumption is seen in many prior works\cite{chakraborty2019software,Chakraborty2021BiasIM,chen2022maat,peng2022fairmask}, especially those in which data pre-processing methods are proposed to mitigate bias from training data. However, few studies comment on the second assumption, relying instead on some simplistic randomized selection procedure to create the train/test tests. Hence,  (a) training data may under-represents certain protected social groups (e.g., gender, race, age), (b) when testing the model on data of the same distribution,  protected groups will be discriminated against, and (c) the same distribution is expected in future upcoming data.
While we agree to parts (a) and (b), we take issue with part (c).
What if biases in training data are due to causes (e.g., bad legacy) unrepeatable in the future? Hence we say that fairness testing should not only be conducted on the raw testing data
but on data selected by {\em propensity}.

\begin{figure*}[!t]
\centering
\includegraphics[width=.8\textwidth]{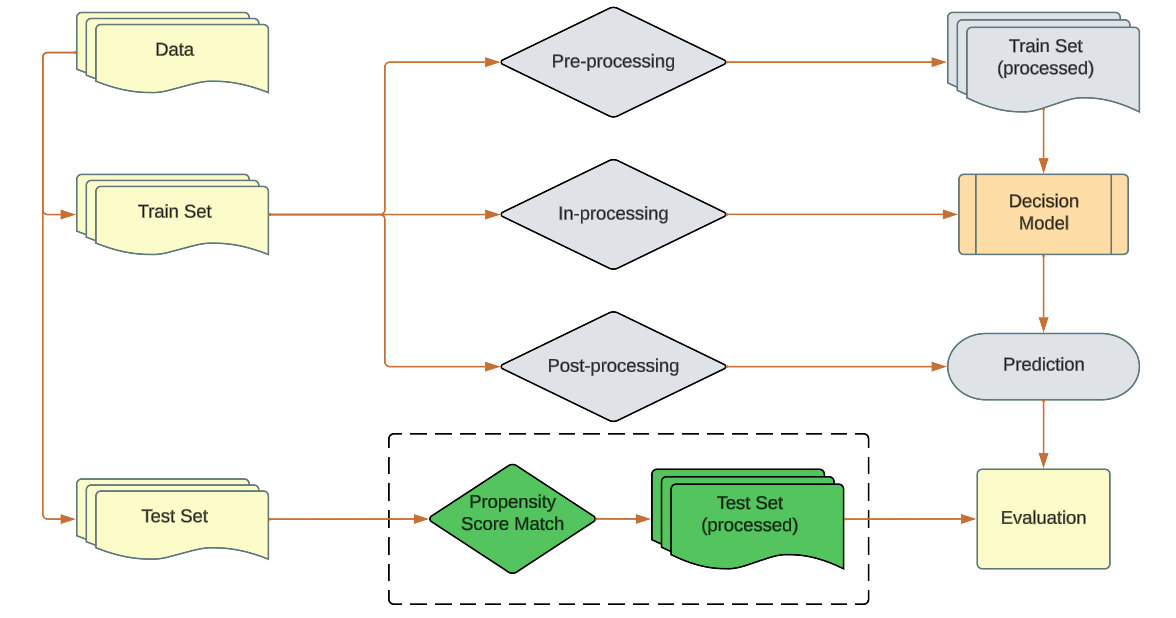}
\caption{Fairness testing proposed in this paper. To evaluate the effectiveness of any de-biasing model, we want to ensure that the test set reflects the ideal data distribution in the future.}
% \caption{Propensity score matching selects samples that not only represent the future distribution of class labels but also represent the (expected) future distribution of protected attributes. }
\label{PSM}\centering
\end{figure*}

Propensity, a concept from the causal reasoning literature,  measures how much a variable affects the outcome (under the assumption that all other inputs are not changed) \cite{rosenbaum1983central}. 
Our analysis of propensity lets us make four contributions:
\begin{itemize}
    \item {\bf Contribution 1}: We show that propensity changes what data should be used in the test set for fairness studies.
    \item {\bf Contribution 2}: We warn that much prior work on fairness, which ignored issues of propensity, needs to be revisited since that work may have used the wrong test data. 
    \item {\bf Contribution 3}: We proposed, in Figure \ref{PSM}, a new schema for fairness testing leveraging propensity score matching, which retired a common threat to validity shared by much prior work (for full details on this scheme, see \S\ref{proposed_approach}).
    \item {\bf Contribution 4}: Using propensity, we design 
    (in \S\ref{proposed_approach}) an adaptive post-processing method for mitigating bias, namely {\bf FairMatch}. Experimentally, we show that our method can achieve on-par or superior performance when compared against prior work (measured in terms of superior fairness/performance trade-offs).
\end{itemize}
The rest of this paper is structured as follows. \S\ref{background} provides fundamental knowledge and related works in the field of fairness testing. \S\ref{methodology} introduces methodologies and preliminary discoveries that motivate our attempts to improve fairness testing as well as bias mitigation. \S\ref{proposed_approach} describes the novel approaches proposed in this paper. \S\ref{experiment} introduces the experiment setup used to evaluate our approach along with other baselines. \S\ref{result} shows experiment results as supportive evidence to our conclusive arguments. 
\S\ref{threat} lists external and internal threats to validity in this paper. Finally, \S\ref{conclusion} presents our conclusions and prospects for future work. 
To better support open science, all scripts and data used in this study are available online at \url{https://github.com/anonymous12138/Fairness-PSM}.

\section{Background}\label{background}
This section presents some fundamental concepts of software fairness, including fairness metrics and related works attempting to achieve better fairness.
% \section{Problem Description}\label{rw}
% In this section, we introduce (a) fundamental concepts of  software fairness and 
% (b) related work that tries to ensure it. 

\subsection{Why Study Software Fairness?}\label{whyfairness}

Sometimes, we are asked, ``is fairness testing really a software engineering problem?''. In reply to that question, we answer the following. 
% Fairness testing is an integral component of the software engineering lifecycle, addressing critical ethical considerations and societal implications associated with the deployment of software systems. 
As software engineers design, develop, and deploy applications that impact diverse user groups, ensuring fairness becomes paramount to mitigate the risk of unintentional biases and discriminatory outcomes. This intersection of fairness testing and software engineering emphasizes the commitment to developing inclusive and equitable technologies, aligning with the ethical responsibilities that software engineers bear in the creation and maintenance of software systems. The incorporation of fairness testing in the software engineering workflow not only fosters the development of socially responsible applications but also contributes to the establishment of best practices for addressing the many examples fairness concerns seen in current systems:

\begin{itemize}
\item {\bf Donation proposal:} Proposals from low-income groups are far   more  likely to be incorrectly ignored by donation groups~\cite{cruz2021promoting};
\item {\bf Income estimation:} Females can be five times more likely to be incorrectly classified as low-income~\cite{ADULT};
\item {\bf Parole assessment:} African Americans are five times more likely to be denied bail, then languish in prison until trial~\cite{COMPAS}.  
\item {\bf Facial recognition:} It is reported that commercial facial recognition systems are exposed to gender and skin-type bias~\cite{buolamwini2018gender}, while other research investigates the unregulated use of facial recognition technology in law enforcement~\cite{garvie2016perpetual}.
\item {\bf Large Language Models:} Large language models (LLMs) may inadvertently capture and perpetuate societal biases present in training data when their embeddings reflect and amplify various biases, including gender, race, and societal stereotypes~\cite{caliskan2017semantics}. 
\end{itemize}
These are are many such examples\cite{rudin2019explaining}\footnote{See also 
\url{http://tiny.cc/bad23a}, \url{http://tiny.cc/bad23b}, \url{http://tiny.cc/bad23c}}.
For   example, the last chapter of Noble~\cite{noble2018algorithms} describes how a successful hair salon went bankrupt due to internal choices within the YELP recommendation algorithm.

Mathews~\cite{mathews23} argues that everyone seeks ways to exploit some advantage for themselves. Hence, we should expect the software we build to discriminate against some social groups;
\begin{quote}
{\em 
    ``People often think of their own hard work or a good decision they made. However, it is often more accurate to look at advantages like the ability to borrow money from family and friends when you are in trouble, deep network connections so that you hear about opportunities or have a human look at your application,  the ability to move on from a mistake that might send someone else to jail...... Success often comes from exploiting a playing field that is far from level and when push comes to shove, we often want those advantages for our children, our family, our friends, our community, our organizations.''}
 \end{quote}
% Hence, I assert that unfairness is a widespread issue that needs to be addressed and managed. Specifically, we need to ensure that a software system created by one group, $A$, can be critiqued and modified by another group, $B$.
To address these issues, software development organizations should review their hiring practices to diversify the range of perspectives seen in design teams.  Requirements engineering practices should be improved to include extensive communication with the stakeholders of the software. 
 Software testing teams should extend their tests to cover issues such as discrimination against specific social groups~\cite{cruz2021promoting,10.1145/3585006,Chakraborty2021BiasIM}.

\begin{table*}[t!]
\centering
% \small
\tiny
% \footnotesize
\caption{Description of datasets used in this paper.}
\begin{tabular}{ccclcc}
\toprule
Dataset &
  \#Features &
  \#Rows &
  \multicolumn{1}{c}{Domain} &
  Protected Attribute &
  Favorable Label \\
  \midrule
Adult Census~\cite{ADULT} & 14  & 48,842 & U.S. census information from 1994 to predict personal income & Sex, Race &
  Income $>$ \$50,000 \\
Compas~\cite{COMPAS}         & 28    & 7,214  & Criminal history of defendants to predict reoffending & Sex, Race & Re-offend $=$ false   \\
German Credit~\cite{GERMAN}  & 20    & 1,000  & Personal information to predict good or bad credit                              & Sex       & Credit $=$ good   \\
Bank Marketing~\cite{BANK} & 16 & 45,211 & Marketing data of a Portuguese bank to predict term deposit                     & Age       & Subscription $=$ yes \\
Heart Health~\cite{HEART}   & 14    & 297    & Patient information from Cleveland DB to predict heart disease                  & Age       & Diagnose $=$ yes   \\
% Default Credit~\cite{DEFAULT} & 23   & 30,000 & Customer information in Taiwan to predict default payment         & Sex       & Payment $=$ yes   \\
MEPS15~\cite{MEPS}        & 1831  & 4,870 & Surveys of household members and their medical providers & Race      & Utilization $>=$ 10  \\
\bottomrule
\end{tabular}
\label{tab:dataset}
\end{table*}

\begin{table*}[t!]
% \footnotesize
\centering
% \small
\tiny
\caption{Definitions and descriptions of fairness metrics used in this paper.}

\begin{tabular}{|l|l|l|}
\hline
\multicolumn{1}{c}{Metric} &
  \multicolumn{1}{c}{Definition} &
 \multicolumn{1}{c}{Description} \\ \hline
Average Odds Difference (AOD) &
  \begin{tabular}[c]{@{}l@{}}TPR $=$ TP$/$(TP $+$ FN), FPR $=$ FP/(FP $+$ TN)\\ AOD= (($FPR_{U} - FPR_{P}$) + ($TPR_{U} - TPR_{P}$))$/$2\end{tabular} &
  \begin{tabular}[c]{@{}l@{}}Average of difference in False Positive Rates(FPR) and True\\ Positive Rates(TPR) for unprivileged and privileged groups\end{tabular} \\ \hline
Equal Opportunity Difference (EOD) &
  EOD $=$ $TPR_{U} - TPR_{P}$ &
  \begin{tabular}[c]{@{}l@{}}Difference of True Positive Rates(TPR) for unprivileged and\\ privileged groups\end{tabular} \\ \hline
Statistical Parity Difference (SPD) &
  SPD $=$ P (Y $=$ 1$|$PA $=$ 0) $-$ P (Y $=$ 1$|$PA $=$ 1) &
  \begin{tabular}[c]{@{}l@{}}Difference between probability of unprivileged group \\ (protected attribute PA $=$ 0) gets favorable prediction (Y $=$ 1)\\ \& probability of privileged group (protected attribute PA $=$ 1)\\ gets favorable prediction (Y $=$ 1)\end{tabular} \\ \hline
Disparate Impact (DI) &
  DI $=$ P {(}Y $=$ 1$|$PA $=$ 0{]}$/$P {[}Y $=$ 1$|$PA $=$ 1{)} &
  Similar to SPD but measuring ratio rather than the probability \\ \hline
% Flip Rate (FR) &
%   FR $=$ $\Sigma$(L$|$L{[}PA$=$0{]} $\not =$ L{[}PA$=$1{]})$/ total$ &
%   \begin{tabular}[c]{@{}l@{}}The ratio of instances whose predicted label ($L$) will change\\ when flipping their protected attributes (e.g., PA$=$1 to PA$=$0) \end{tabular} \\ \hline
\end{tabular}
\label{tab:metrics}
\end{table*}

Further, on the legal front,
Canellas~\cite{canellas21} and Mathews et al.~\cite{matthewsshould} suggest a tiered process in which the most potentially discriminatory projects are routinely reviewed by an independent external review team
(as done in the IEEE 1012 independent V\&V standard). Ben Green~\cite{green2022flaws} notes that reviewing software systems and AI systems is becoming a legislative necessity and that human-in-the-loop auditing of decisions made by software is often mandatory.
Such legislation is necessary to move away from the internal application of voluntary industrial standards (since, as seen in the Volkswagen emissions scandal (see \url{http://tiny.cc/scandalvw}), companies cannot always be trusted
to voluntarily apply reasonable standards.

We note that our colleagues in AI are also studying algorithmic fairness.
Despite their best efforts, all the above problems are open and pressing. So we argue that fairness is an ``all hand on deck'' situation where anyone with a useful perspective or novel findings
should contribute to this line of research. Hence, this paper.

\subsection{Related Works}
This section provides a more comprehensive context of research works studying software fairness. More specifically,  we illustrate the in-depth connection between software engineering and fairness-aware ML/AI applications.
{\bf SE for fairness}: There has been increasing interest in leveraging software engineering techniques to enhance fairness in machine learning models, with approaches focusing on hyperparameter optimization, ensemble methods, and algorithmic fairness constraints.
\begin{itemize}
    \item {\bf Fairness Repairing and Debugging}:
Chakraborty et al. \cite{chakraborty2019software} were among the first to advocate for incorporating fairness as a primary goal during hyperparameter optimization, demonstrating that this approach can reduce discrimination while maintaining predictive accuracy. Building on this foundation, Tizpaz-Niari et al. \cite{tizpaz2022fairness} explored how specific hyperparameter configurations can significantly improve fairness without sacrificing precision, using search-based software testing to identify the fairness-precision frontier. Similarly, Linear-regression-based Training Data Debugging (LTDD) \cite{li2022training} was introduced, which focuses on identifying and removing biased features in training data, refining fairness in machine learning models with minimal performance impact. RUNNER \cite{li2024runner} further enhances fairness in deep neural networks by efficiently diagnosing and repairing unfair neurons, reducing computational overhead while improving fairness across different datasets.

\item {\bf Ensemble and Search-based Optimization}:
MAAT \cite{chen2022maat} presented an innovative approach that optimizes the trade-off between fairness and performance by combining models with distinct objectives, effectively addressing fairness bugs without compromising accuracy. Likewise, Gohar \cite{gohar2023towards} demonstrated how careful design and configuration of ensemble models could achieve fairer outcomes without the need for additional bias mitigation techniques. Recently, a novel search-based method for repairing fairness issues in decision-making software \cite{hort2024search} was introduced, achieving simultaneous improvements in both fairness and accuracy. Lastly, MirrorFair \cite{xiao2024mirrorfair} leveraged counterfactual predictions within an adaptive ensemble approach to balance fairness and performance, illustrating how search-based configurations and ensemble techniques can be effectively applied to diverse decision-making tasks.

    \item {\bf Requirement Engineering for Software Fairness}:
The Seldonian Toolkit \cite{hoag2023seldonian} provides software engineers with the ability to integrate domain-specific fairness and safety requirements into machine learning systems, ensuring these requirements are met through provably safe and fair algorithms. This toolkit is particularly effective in high-stakes domains such as healthcare and criminal justice, where fairness is critical. A recent survey \cite{ferrara2024fairness} found that fairness is often a secondary concern in AI development, with practitioners lacking the understanding and tools to handle it, highlighting the need to address fairness throughout the software lifecycle.  Additionally, a comprehensive survey \cite{chen2024fairness} highlights the challenge of designing automatic techniques for constructing reliable oracles for fairness testing. This underscores the importance of requirement engineering in establishing clear and accepted fairness criteria, facilitating more effective testing and mitigation strategies in software systems.
\end{itemize}
{\bf Fairness in SE}: As software systems increasingly play a critical role in societal decision-making processes, ensuring fairness in their development and operation has become a key focus in software engineering.
\begin{itemize}
    \item {\bf Bias Detection and Mitigation}:
Bias detection and mitigation have become a focus of efforts across various fields closely connected to SE. Zhang et al. \cite{zhang2021ignorance} explore how feature sets and training data impact fairness in machine learning, revealing that larger feature sets can enhance both fairness and accuracy, whereas increasing training data without ensuring diversity may lead to reduced fairness. In addressing the need for fairness in AI software, Fairway \cite{Chakraborty_2020} proposes a combined pre-processing and in-processing approach to detect and mitigate ethical bias in machine learning models. The authors emphasize that bias testing and mitigation should be integral to the software development lifecycle, demonstrating that bias can be reduced without significantly compromising predictive performance. Zhang et al. \cite{zhang2020white} introduce a scalable approach for detecting individual discriminatory instances in deep neural networks (DNNs) using adversarial sampling. By leveraging efficient techniques like gradient computation and clustering, this method improves bias detection in DNNs, making fairness testing more effective and less time-consuming in socially impactful applications. A fairness-aware recommendation system, iRec2.0 \cite{wang2022context}, is reported to improve crowdworker allocation in crowdtesting tasks by dynamically adjusting recommendations to reduce popularity bias, integrating fairness into software testing practices.

\item {\bf Empirical Studies}:
Chen et al. \cite{chen2023comprehensive} provide an empirical analysis of 17 bias mitigation methods, highlighting the difficulty in achieving an optimal trade-off between fairness and performance, further informing fairness improvements in production systems. The study points out that the effectiveness of these methods is highly dependent on the context, differing across tasks, models, and selected protected attributes, which strongly supports the need for emerging empirical research on specific topics. Another comprehensive survey of fairness testing in machine learning software \cite{chen2024fairness}, categorized research by testing workflows and components, and identified key datasets and tools instrumental in measuring and detecting bias. Yang et al. \cite{yang2024large} present the first large-scale empirical comparison of 13 fairness-improving methods in image classification, revealing significant performance variations and recommending pre-processing techniques as the most effective for enhancing fairness. Fairea \cite{hort2021fairea}, a benchmarking framework that evaluates the effectiveness of machine learning bias mitigation methods using a model behavior mutation approach, is utilized in a large-scale empirical study and made publicly available to aid researchers and software engineers in improving bias mitigation strategies in ML systems.

\item {\bf Policy and Regulations in SE community}:
The White House's Blueprint for an AI Bill of Rights \cite{whitehouse2022blueprint} highlights concerns over technology and automated systems that threaten civil rights by perpetuating bias and discrimination, urging software engineers to root out inequity and embed fairness in decision-making processes. Similarly, the European Union's Artificial Intelligence Act \cite{EU2024AIA} outlines principles including diversity, non-discrimination, and fairness, mandating that high-risk AI systems maintain appropriate levels of accuracy and robustness. A Joint Statement by U.S. Federal Agencies \cite{JointStatement2024} reaffirms America's commitment to fairness and equality, emphasizing the need for responsible innovation to abide by civil rights legislation and the existence of standards from organizations like the Department of Homeland Security that prevent AI from making judgments based on inappropriate criteria like gender or race.

\end{itemize}

\subsection{Problem Description}
Software systems assisted by ML models play important roles in many domains. Fairness in ML software refers to the impartial treatment of individuals of any social group (as defined by age, gender, race, etc.). In this paper, we stress binary classification tasks where:  
\begin{itemize}
    \item A {\em favorable label} is the label that provides an individual with benefits such as being estimated with a low risk of re-offending in parole assessments. 
    \item A {\em protected/sensitive attribute} indicates the social groups to which data instances belong, such as gender, race, and age. 
    \item Based on the protected attribute, a subgroup of individuals is {\em privileged} if they are more likely to receive the favorable label over the {\em unprivileged} group. 
\end{itemize}
Table~\ref{tab:dataset} shows five fairness datasets used in this paper, some of which possess more than one protected attribute. Motivated by the demand to ensure fairness, many metrics have been defined to quantify the level of fairness/discrimination of ML models.

% Narayanan~\cite{Arvind} has defined 21 different versions of fairness.
% Based on prior literature~\cite{Biswas_2020, Chakraborty_2020, Chakraborty2021BiasIM}, among these 21 versions, two specific versions of fairness are widely explored and given the greatest importance. 
% We decided to explore the same two versions and chose different metrics to evaluate them. 

\begin{itemize}
    \item {\em Group fairness} requires the approximate equalization
    of certain statistical properties across groups divided by the protected attribute. In this paper, we use 4 group fairness metrics that were widely used in previous research~\cite{kamiran2012data, feldman2015certifying, Chakraborty_2020, Chakraborty2021BiasIM, Biswas_2020}.
    \item {\em Individual fairness} is concerned with the treatment of similar individuals. The principle of individual fairness posits that individuals who are similar in relevant features should receive similar outcomes from an ML model, regardless of their belonging to different demographic groups. 
\end{itemize}
It is tricky to define "similarity" between individuals, especially with the growth of amount of features used in a task. The curse of dimensionality made it difficult to determine the level of similarity upon which two individuals shall expect to receive similar treatments. Therefore, it is not surprising that many prior works focus on group fairness.
Table~\ref{tab:metrics} contains mathematical definitions of four fairness metrics used in our experiment evaluation. All the fairness metrics are calculated based on the binary classification confusion matrix, which consists of four parts: true positive (TP), true negative (TN), false positive (FP), and false negative (FN). Notably, although we only use group fairness in this paper, our proposed approach is greatly motivated by the notion that comparable individuals ought to receive similar model outcomes.

\begin{table*}[]
\centering
\small
\caption{The performance and fairness scores measured from testing samples. The test set is split into two subsets depending on the corresponding sampling strategy specified in \S\ref{sampling_strategy}. The ratio column shows, in percentage, the portion of each subset.}
\label{tab:PSM-subgroup}
\resizebox{\textwidth}{!}{%
\begin{tabular}{|l|l|l|cccc|cccc|}
\hline
Dataset                      & Method      & Ratio & Accuracy & Precision & Recall & F1 & AOD & EOD & SPD & DI \\ \hline
\multirow{5}{*}{Adult: Sex}   & PSM\_sampled & 0.19  & 91       & 74        & 50     & 59 & 2   & 3   & 0   & 0  \\
                             & C\_sampled   & 0.67  & 88       & 49        & 45     & 47 & 8   & 26  & 11  & 68 \\
                             & PA\_sampled  & 0.49  & 71       & 85        & 52     & 64 & 6   & 23  & 16  & 50 \\
                             & WAE\_sampled & 0.15  & 69       & 89        & 43     & 58 & 7   & 21  & 14  & 45 \\
                             & Original    & 1.00  & 83       & 71        & 52     & 60 & 8   & 23  & 19  & 78 \\ \hline
\multirow{5}{*}{Adult: Race}  & PSM\_sampled & 0.19  & 89       & 71        & 48     & 57 & 0   & 1   & 0   & 0  \\
                             & C\_sampled   & 0.30  & 88       & 66        & 53     & 59 & 6   & 16  & 5   & 31 \\
                             & PA\_sampled  & 0.49  & 72       & 89        & 52     & 65 & 2   & 7   & 5   & 16 \\
                             & WAE\_sampled & 0.10  & 69       & 89        & 43     & 58 & 7   & 21  & 14  & 45 \\
                             & Original    & 1.00  & 83       & 71        & 52     & 60 & 2   & 7   & 8   & 44 \\ \hline
\end{tabular}
}
\end{table*}

\newcommand{\YES}{CCE5D2}   %7CC444
\newcommand{\NO}{EED9D9} % FD6864

\begin{table*}[]
\centering
\small
\caption{The full table of comparisons listed in Table \ref{tab:PSM-subgroup}. Each cell represents the change in performance/fairness metrics between the original test data and data selected by different sampling methods in \S\ref{sampling_strategy}. An increase in performance metrics is preferred, and so is a decrease in fairness metrics. Better/worse metric values are marked in green/red. }
\label{tab:subgroup2}
\resizebox{\textwidth}{!}{%
\begin{tabular}{|l|l|cccc|cccc|}
\hline
Dataset &
  Method &
  Acc &
  Precision &
  Recall &
  F1 &
  aod &
  eod &
  spd &
  di \\ \hline
 &
  PSM\_sampled &
  \cellcolor[HTML]{\YES}0.10 &
  0.04 &
  -0.04 &
  -0.02 &
  \cellcolor[HTML]{\YES}-0.75 &
  \cellcolor[HTML]{\YES}-0.87 &
  \cellcolor[HTML]{\YES}-1.00 &
  \cellcolor[HTML]{\YES}-1.00 \\
 &
  C\_sampled &
  \cellcolor[HTML]{\YES}0.06 &
  \cellcolor[HTML]{\NO}-0.31 &
  \cellcolor[HTML]{\NO}-0.13 &
  \cellcolor[HTML]{\NO}-0.22 &
  0.00 &
  \cellcolor[HTML]{\NO}0.13 &
  \cellcolor[HTML]{\YES}-0.42 &
  \cellcolor[HTML]{\YES}-0.13 \\
 &
  PA\_sampled &
  \cellcolor[HTML]{\NO}-0.14 &
  \cellcolor[HTML]{\YES}0.20 &
  0.00 &
  \cellcolor[HTML]{\YES}0.07 &
  \cellcolor[HTML]{\YES}-0.25 &
  0.00 &
  \cellcolor[HTML]{\YES}-0.16 &
  \cellcolor[HTML]{\YES}-0.36 \\
\multirow{-4}{*}{Adult: Sex} &
  WAE\_sampled &
  \cellcolor[HTML]{\NO}-0.17 &
  \cellcolor[HTML]{\YES}0.25 &
  \cellcolor[HTML]{\NO}-0.17 &
  -0.03 &
  \cellcolor[HTML]{\YES}-0.13 &
  \cellcolor[HTML]{\YES}-0.09 &
  \cellcolor[HTML]{\YES}-0.26 &
  \cellcolor[HTML]{\YES}-0.42 \\ \hline
 &
  PSM\_sampled &
  \cellcolor[HTML]{\YES}0.07 &
  0.00 &
  \cellcolor[HTML]{\NO}-0.08 &
  \cellcolor[HTML]{\NO}-0.05 &
  \cellcolor[HTML]{\YES}-1.00 &
  \cellcolor[HTML]{\YES}-0.86 &
  \cellcolor[HTML]{\YES}-1.00 &
  \cellcolor[HTML]{\YES}-1.00 \\
 &
  C\_sampled &
  \cellcolor[HTML]{\YES}0.06 &
  \cellcolor[HTML]{\NO}-0.07 &
  0.02 &
  -0.02 &
  \cellcolor[HTML]{\NO}2.00 &
  \cellcolor[HTML]{\NO}1.29 &
  \cellcolor[HTML]{\YES}-0.38 &
  \cellcolor[HTML]{\YES}-0.30 \\
 &
  PA\_sampled &
  \cellcolor[HTML]{\NO}-0.13 &
  \cellcolor[HTML]{\YES}0.25 &
  0.00 &
  \cellcolor[HTML]{\YES}0.08 &
  0.00 &
  0.00 &
  \cellcolor[HTML]{\YES}-0.38 &
  \cellcolor[HTML]{\YES}-0.64 \\
\multirow{-4}{*}{Adult: Race} &
  WAE\_sampled &
  \cellcolor[HTML]{\NO}-0.17 &
  \cellcolor[HTML]{\YES}0.25 &
  \cellcolor[HTML]{\NO}-0.17 &
  -0.03 &
  \cellcolor[HTML]{\NO}2.50 &
  \cellcolor[HTML]{\NO}2.00 &
  \cellcolor[HTML]{\NO}0.75 &
  0.02 \\ \hline
 &
  PSM\_sampled &
  \cellcolor[HTML]{\YES}0.05 &
  0.03 &
  \cellcolor[HTML]{\YES}0.19 &
  \cellcolor[HTML]{\YES}0.10 &
  \cellcolor[HTML]{\NO}{\color[HTML]{000000} 0.50} &
  \cellcolor[HTML]{\YES}-0.83 &
  \cellcolor[HTML]{\YES}-1.00 &
  \cellcolor[HTML]{\YES}-1.00 \\
 &
  C\_sampled &
  0.02 &
  \cellcolor[HTML]{\YES}0.15 &
  0.00 &
  \cellcolor[HTML]{\YES}0.07 &
  \cellcolor[HTML]{\YES}-0.50 &
  \cellcolor[HTML]{\NO}{\color[HTML]{000000} 0.33} &
  \cellcolor[HTML]{\YES}-0.27 &
  \cellcolor[HTML]{\YES}-0.31 \\
 &
  PA\_sampled &
  0.00 &
  \cellcolor[HTML]{\NO}{\color[HTML]{000000} -0.06} &
  0.00 &
  -0.03 &
  0.00 &
  \cellcolor[HTML]{\NO}{\color[HTML]{000000} 1.33} &
  \cellcolor[HTML]{\NO}{\color[HTML]{000000} 0.09} &
  \cellcolor[HTML]{\NO}{\color[HTML]{000000} 0.06} \\
\multirow{-4}{*}{Compas: Sex} &
  WAE\_sampled &
  0.00 &
  \cellcolor[HTML]{\NO}{\color[HTML]{000000} -0.08} &
  \cellcolor[HTML]{\YES}0.07 &
  0.00 &
  0.00 &
  0.00 &
  \cellcolor[HTML]{\YES}-0.18 &
  \cellcolor[HTML]{\YES}-0.19 \\ \hline
 &
  PSM\_sampled &
  0.00 &
  \cellcolor[HTML]{\YES}0.05 &
  \cellcolor[HTML]{\YES}0.13 &
  \cellcolor[HTML]{\YES}0.07 &
  \cellcolor[HTML]{\YES}{-0.17} &
  \cellcolor[HTML]{\YES}-0.56 &
  \cellcolor[HTML]{\YES}-1.00 &
  \cellcolor[HTML]{\YES}-1.00 \\
 &
  C\_sampled &
  -0.03 &
  \cellcolor[HTML]{\YES}0.05 &
  0.01 &
  \cellcolor[HTML]{\YES}0.03 &
  \cellcolor[HTML]{\YES}{ -0.67} &
  \cellcolor[HTML]{\NO}{\color[HTML]{000000} 0.11} &
  \cellcolor[HTML]{\YES}-0.20 &
  \cellcolor[HTML]{\YES}-0.25 \\
 &
  PA\_sampled &
  \cellcolor[HTML]{\NO}-0.05 &
  \cellcolor[HTML]{\NO}-0.09 &
  -0.03 &
  \cellcolor[HTML]{\NO}{\color[HTML]{000000} -0.06} &
  \cellcolor[HTML]{\YES}{ -0.83} &
  \cellcolor[HTML]{\NO}{\color[HTML]{000000} 0.11} &
  \cellcolor[HTML]{\YES}-0.10 &
  \cellcolor[HTML]{\YES}-0.13 \\
\multirow{-4}{*}{Compas: Race} &
  WAE\_sampled &
  \cellcolor[HTML]{\NO}-0.05 &
  \cellcolor[HTML]{\NO}-0.09 &
  0.03 &
  -0.03 &
  \cellcolor[HTML]{\YES}{ -0.83} &
  \cellcolor[HTML]{\YES}-0.22 &
  \cellcolor[HTML]{\YES}-0.40 &
  \cellcolor[HTML]{\YES}-0.38 \\ \hline
 &
  PSM\_sampled &
  0.00 &
  0.00 &
  \cellcolor[HTML]{\NO}-0.07 &
  -0.03 &
  \cellcolor[HTML]{\YES}-0.20 &
  \cellcolor[HTML]{\YES}-0.57 &
  \cellcolor[HTML]{\YES}-1.00 &
  \cellcolor[HTML]{\YES}-1.00 \\
 &
  C\_sampled &
  0.01 &
  0.00 &
  0.00 &
  0.00 &
  \cellcolor[HTML]{\YES}-1.00 &
  \cellcolor[HTML]{\NO}0.43 &
  0.00 &
  0.00 \\
 &
  PA\_sampled &
  \cellcolor[HTML]{\NO}-0.25 &
  \cellcolor[HTML]{\NO}-0.30 &
  0.00 &
  \cellcolor[HTML]{\NO}-0.20 &
  \cellcolor[HTML]{\YES}-0.20 &
  \cellcolor[HTML]{\NO}0.29 &
  \cellcolor[HTML]{\NO}0.30 &
  \cellcolor[HTML]{\NO}0.30 \\
\multirow{-4}{*}{German: Sex} &
  WAE\_sampled &
  \cellcolor[HTML]{\NO}-0.25 &
  \cellcolor[HTML]{\NO}-0.29 &
  -0.03 &
  \cellcolor[HTML]{\NO}-0.20 &
  \cellcolor[HTML]{\NO}0.20 &
  \cellcolor[HTML]{\YES}-0.14 &
  \cellcolor[HTML]{\NO}0.10 &
  \cellcolor[HTML]{\NO}0.20 \\ \hline
 &
  PSM\_sampled &
  \cellcolor[HTML]{\NO}-0.05 &
  -0.03 &
  \cellcolor[HTML]{\YES}0.09 &
  0.03 &
  \cellcolor[HTML]{\YES}-0.55 &
  \cellcolor[HTML]{\YES}-0.89 &
  \cellcolor[HTML]{\YES}-1.00 &
  \cellcolor[HTML]{\YES}-1.00 \\
 &
  C\_sampled &
  \cellcolor[HTML]{\YES}0.06 &
  \cellcolor[HTML]{\YES}0.23 &
  \cellcolor[HTML]{\YES}0.08 &
  \cellcolor[HTML]{\YES}0.15 &
  \cellcolor[HTML]{\NO}0.09 &
  \cellcolor[HTML]{\YES}-0.56 &
  \cellcolor[HTML]{\YES}-0.91 &
  \cellcolor[HTML]{\YES}-0.96 \\
 &
  PA\_sampled &
  0.00 &
  0.04 &
  0.00 &
  0.02 &
  \cellcolor[HTML]{\YES}-0.09 &
  \cellcolor[HTML]{\YES}-0.33 &
  \cellcolor[HTML]{\YES}-0.83 &
  \cellcolor[HTML]{\YES}-0.86 \\
\multirow{-4}{*}{Bank: Age} &
  WAE\_sampled &
  -0.01 &
  \cellcolor[HTML]{\YES}0.08 &
  \cellcolor[HTML]{\NO}-0.11 &
  -0.02 &
  \cellcolor[HTML]{\NO}1.64 &
  \cellcolor[HTML]{\NO}3.78 &
  \cellcolor[HTML]{\YES}-0.39 &
  \cellcolor[HTML]{\YES}-0.20 \\ \hline
 &
  PSM\_sampled &
  \cellcolor[HTML]{\NO}-0.05 &
  \cellcolor[HTML]{\NO}-0.06 &
  \cellcolor[HTML]{\NO}-0.06 &
  \cellcolor[HTML]{\NO}-0.06 &
  \cellcolor[HTML]{\NO}6.00 &
  \cellcolor[HTML]{\YES}-0.46 &
  \cellcolor[HTML]{\YES}-1.00 &
  \cellcolor[HTML]{\YES}-1.00 \\
 &
  C\_sampled &
  \cellcolor[HTML]{\NO}-0.06 &
  \cellcolor[HTML]{\NO}-0.10 &
  \cellcolor[HTML]{\NO}-0.10 &
  \cellcolor[HTML]{\NO}-0.10 &
  \cellcolor[HTML]{\NO}3.00 &
  \cellcolor[HTML]{\YES}-0.29 &
  \cellcolor[HTML]{\YES}-0.38 &
  \cellcolor[HTML]{\YES}-0.27 \\
 &
  PA\_sampled &
  \cellcolor[HTML]{\NO}-0.06 &
  -0.04 &
  \cellcolor[HTML]{\NO}-0.05 &
  \cellcolor[HTML]{\NO}-0.05 &
  \cellcolor[HTML]{\YES}-1.00 &
  \cellcolor[HTML]{\YES}-0.07 &
  \cellcolor[HTML]{\YES}-0.35 &
  \cellcolor[HTML]{\YES}-0.33 \\
\multirow{-4}{*}{Heart: Age} &
  WAE\_sampled &
  0.00 &
  \cellcolor[HTML]{\YES}0.06 &
  \cellcolor[HTML]{\NO}-0.05 &
  0.01 &
  \cellcolor[HTML]{\YES}-1.00 &
  \cellcolor[HTML]{\NO}0.07 &
  \cellcolor[HTML]{\YES}-0.25 &
  \cellcolor[HTML]{\YES}-0.21 \\ \hline
 &
  PSM\_sampled &
  -0.02 &
  \cellcolor[HTML]{\YES}0.08 &
  \cellcolor[HTML]{\NO}-0.15 &
  \cellcolor[HTML]{\NO}-0.06 &
  \cellcolor[HTML]{\NO}0.67 &
  \cellcolor[HTML]{\NO}0.50 &
  \cellcolor[HTML]{\YES}-1.00 &
  \cellcolor[HTML]{\YES}-1.00 \\
 &
  C\_sampled &
  \cellcolor[HTML]{\YES}0.05 &
  \cellcolor[HTML]{\NO}-0.15 &
  -0.02 &
  \cellcolor[HTML]{\NO}-0.06 &
  \cellcolor[HTML]{\NO}1.00 &
  \cellcolor[HTML]{\NO}0.67 &
  \cellcolor[HTML]{\YES}-1.00 &
  \cellcolor[HTML]{\YES}-0.88 \\
 &
  PA\_sampled &
  \cellcolor[HTML]{\NO}-0.21 &
  \cellcolor[HTML]{\YES}0.52 &
  0.00 &
  \cellcolor[HTML]{\YES}0.16 &
  0.00 &
  0.00 &
  \cellcolor[HTML]{\YES}-0.25 &
  \cellcolor[HTML]{\YES}-0.62 \\
\multirow{-4}{*}{MEPS15: Race} &
  WAE\_sampled &
  \cellcolor[HTML]{\NO}-0.21 &
  \cellcolor[HTML]{\YES}0.42 &
  \cellcolor[HTML]{\YES}0.10 &
  \cellcolor[HTML]{\YES}0.22 &
  \cellcolor[HTML]{\NO}3.00 &
  \cellcolor[HTML]{\NO}2.17 &
  \cellcolor[HTML]{\NO}0.50 &
  \cellcolor[HTML]{\YES}-0.35 \\ \hline
  % {\bf STD} &
  % \multicolumn{1}{c|}{} &
  % \bf 0.09 &
  % \bf 0.18 &
  % \bf 0.08 &
  % \bf 0.09 &
  % \bf 1.57 &
  % \bf 0.94 &
  % \bf 0.46 &
  % \bf 0.42 \\ \hline
\end{tabular}
}
\end{table*}

\subsection{How to Fix Fairness Bugs}\label{bias-mitigation}
In the literature, bias mitigation methods can be categorized into three major groups, depending on when the mitigation procedure is performed.\\
% \begin{itemize}
    % \item
    \noindent
    {\bf Pre-processing}: Pre-processing algorithms practice fixing fairness bugs in ML-assisted software by transforming the training data that the model learns from.  Fair-SMOTE~\cite{Chakraborty2021BiasIM} uses oversampling techniques to synthetic instances among the training data such that distributions between different target labels and different protected attributes can be re-balanced. Reweighing was proposed by Kamiran et al.~\cite{kamiran2012data} to learn a probabilistic threshold that can generate weights for different instances in training samples according to the combination (protected and class attributes).
    \\
    % \item 
    \noindent
    {\bf In-processing}: In-processing methods tend to mitigate bias during the model training phase. The data set is typically divided into three parts: training, validation, and testing. The learner is fitted to the training set and then optimized to the validation set using both performance and fairness metrics as objectives. Kamishima et al.~\cite{10.1007/978-3-642-33486-3_3} developed Prejudice Remover, which adds a discrimination-aware regularization to the learning objective of the prediction model. 
    Similarly, exponentiated gradient reduction (EGR) is another in-processing technique that reduces fair classification to a sequence of cost-sensitive classification problems. Given the specific fairness metrics selected by users, EGR will return a randomized model with the lowest empirical error subject to the corresponding constraints \cite{agarwal2018reductions}. \\
    % MAAT \cite{chen2022maat} is another recent work that optimizes the performance-fairness trade-off via ensemble learning.

    % \item
    \noindent
    {\bf Post-processing}: This approach believes that bias can be removed by identifying and then reversing biased outcomes from the classification model. Without requiring access to the training data or the model itself, post-processing algorithms will identify model outcomes that are likely to be biased and change them accordingly. Reject Option Classification \cite{Kamiran:2018:ERO:3165328.3165686} is an approach that first identifies the decision boundary with the highest uncertainty. Within that region, the method will adjust the ratio between favorable labels on unprivileged groups and unfavorable labels on privileged groups. Fax-AI \cite{grabowicz2022marrying} is another post-processing method that mitigates model bias by eliminating potential proxy discrimination. Fax-AI limits the usage of certain features, as those features are believed to have a greater likelihood of serving as surrogates for protected attributes. 
    \\
% \end{itemize}

In particular, the above three genres are defined based on the phase in which an algorithm intervenes to mitigate bias. There are also many methods that cannot be explicitly categorized into a single genre. For example, the FairMask framework proposed by Peng et al. \cite{peng2022fairmask} generates synthesized protected attributes (PAs) parallel to the model training phase, and synthetic attributes are used to replace real protected attributes during inference time. FairMask could be considered a post-processing method since it mitigates bias by identifying and reversing biased model outcomes via the use of synthetic PAs. However, the synthesis procedure can occur before/during the model-training phase. Therefore, it is difficult to categorize this algorithm into any of the three types. Another example is MAAT \cite{chen2022maat}. As one of the most recent ensemble-based fairness learning methods, MAAT has outperformed several previous SOTA methods, including Fairway and Fair-SMOTE, by obtaining a better performance-fairness trade-off in many studied datasets. The algorithm of MAAT contains both data re-sampling (pre-processing) and training multiple models for different objectives (in-processing).

As later discussed in \S\ref{experiment}, we select some of the bias mitigation methods mentioned above in each category as the benchmark methods in our experimentation.

\section{Observation}\label{methodology}
In this section, we demonstrate and analyze observations found in the traditional procedure of fairness testing. After that, we introduce techniques based on which we attempt to improve the testing procedure. 
\subsection{Fairness Testing}
Fairness testing refers to a specific aspect of evaluating machine learning models. It usually involves assessing the consistency of a model's impact on different subgroups of the population (identified by their personal attributes). In ML-assisted software, fairness testing plays a crucial role as it ensures that the released software provides equitable outcomes regardless of demographic or other characteristic factors. 

As summarized in the literature, fairness testing can be performed in various components of the ML pipeline: data testing aims to ensure the training data is free of bias-introducing features; ML program testing aims to assess whether the procedures of preparing an ML model, such as feature engineering and hyperparameter tuning may introduce unintentional bias; Model testing such as black-box and white-box testing aims to evaluate the final outcomes of an ML model through certain pre-defined fairness definitions along with corresponding thresholds.

Recent studies have highlighted that fairness testing may be compromised when there are distribution shifts between training and deployment data. Such shifts include demographic shifts \cite{giguere2022fairness}, distribution shifts \cite{an2022transferring}, correlation shifts \cite{Roh2023Improving}, and distribution shifts in graph-structured data \cite{Li2024Graph}, which can lead to models exhibiting unfairness in deployment despite appearing fair during testing.

In this paper, when we discuss fairness testing, we specifically refer to model testing, which examines the distributive fairness of model outcomes on the testing set. 
Moreover, we extrapolate a quite different dimension of fairness testing: the level of trust for fairness testing. We argue that the existing fairness testing may be flawed when the randomly selected testing data do not reflect the true distribution of the population. In other words, fairness testing may sometimes provide plausible yet misinformed insights to practitioners. 

\begin{algorithm}[t!]
\caption{Propensity Score Matching (PSM) \label{algo1}}
% \small
\KwData{$PropensityModel$ the model (usually logistic regression) used to estimate propensity score; $X_{train}$, $y_{train}$ the training data; $X_{test}$, $y_{test}$ the test data; $PA$ the protected attribute 
}
\KwResult{The test set is returned in two subgroups: the one containing samples paired by PSM, and the other one containing the rest of test data.}
\Begin{
    $matched \gets$ $\emptyset$\\
    $not\_matched \gets$ $\emptyset$\\
    $PropensityModel.fit(X_{train}, y_{train})$ \\
    $ps \gets PropensityModel.predict(X_{test})$\\
    % $knn.initialize()$\\
    % $knn.fit(ps)$\\
    $neighbors \gets \{\}$\\
    \ForEach{$x_{test}$ \textbf{in} $X_{test}$}{
    $distances \gets \{\}$\\
    \ForEach{$x_{train}$ \textbf{in} $X_{train}$}{
        $d \gets$ $dist(x_{test}, x_{train})$ {// Calculate Euclidean distance}
        $distances.append(d, y_{train})$
    }
    $sorted\_distances \gets$ $distances.sort()$\\
    $k\_neighbors \gets sorted\_distances[:k]$ \\
    $neighbors.append(X_test, k\_neighbors)$ \\
    }
    \While{$X_{test}.size \neq 0$}{
        $row \gets X_{test}$.pop()\\
        \If{$row.PA==1$}{
            \While{$neighbors.size \neq 0$}{
                $neighbor \gets neighbors.get(row)$\\
                \If{$neighbor.PA \neq row.PA$}{
                $matched \gets matched \cup row$\\
                $not\_matched \gets not\_matched \cup neighbor$\\
                $X_{test}.pop(neighbor)$\\
                break \\
                }
            }
        }
    }
    return $matched, not\_matched$
}
\end{algorithm}

\begin{figure*}[!b]
        \centering
        \begin{subfigure}[b]{0.32\textwidth}
            \centering
            \includegraphics[width=\textwidth]{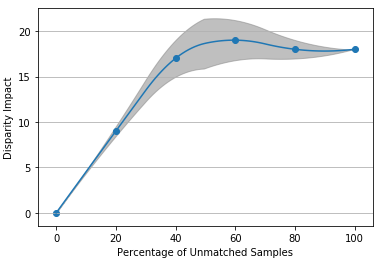}
            \caption[]%
            {{\small German dataset}}    
            \label{fig:german}
        \end{subfigure}
        \hfill
        \begin{subfigure}[b]{0.32\textwidth}
            \centering
            \includegraphics[width=\textwidth]{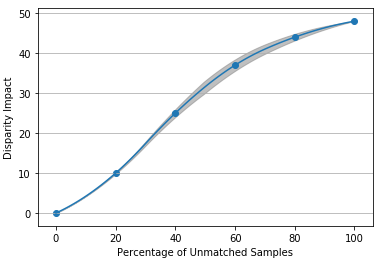}
            \caption[]%
            {{\small Heart dataset}}    
            \label{fig:heart}
        \end{subfigure}
        \hfill
        \begin{subfigure}[b]{0.32\textwidth}  
            \centering 
            \includegraphics[width=\textwidth]{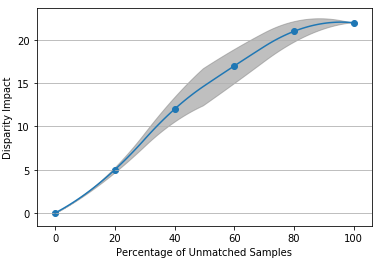}
            \caption[]%
            {{\small MEPS dataset}}    
            \label{fig:meps}
        \end{subfigure}
        % \vskip\baselineskip
        \caption[]
        {\small Curves drawn from three datasets. The gray error area indicates the standard deviation as we use different random seeds for sub-sampling. The x-axis represents the percentage of unmatched testing samples used when calculating the fairness scores. The y-axis represents the fairness metric (DI in this case), where lower scores indicate better fairness.} 
        \label{fig:curve}
    \end{figure*}

\subsection{PSM = Propensity Score Matching}\label{sampling_strategy}
As previously discussed in the introduction session, we argue that there exists inconsistency of presumptions in fairness testing. In many bias mitigation works, especially those that focus on pre-processing methods, it is widely believed that certain social subgroups are misrepresented in the current data. This leads to our question: How to obtain the appropriate test set, given the training data is believed to contain bias? 
Our proposed solution in this paper is propensity score matching.

Propensity, a concept we take from the causal reasoning literature,  measures how much a variable affects the outcome (assuming that all other inputs are not changed) \cite{rosenbaum1983central}.
More precisely, the propensity score is a statistical tool used in observational studies and quasi-experimental research to estimate the probability of receiving a particular treatment or intervention based on observed covariates. As illustrated in Algorithm \ref{algo1}, propensity score matching (PSM) utilized the propensity scores as the estimated probabilities of receiving a particular treatment based on observed covariates. This matching process aims to emulate a randomized control trial by balancing covariates across treatment and control groups, thereby enabling more accurate and reliable causal inferences. The sampling steps are stated as follows:
\begin{itemize}
    \item Fit a learner using training data.
    \item Generate predicted probabilities on testing data samples.
    \item For each data point $P$, identify neighboring samples via the k-nearest neighbors (KNN) algorithm. 
    \item If $P$ has a nearest neighbor $N$ with an opposite PA, add $P$ and $N$ into the treatment and control group respectively. 
\end{itemize}

To compare the influence of PSM with other techniques, this paper includes three other under-sampling methods:
\begin{itemize}
    \item {\bf Class-based sampling}: This method maintains the balance of favorable and unfavorable labels within test data.
    \item {\bf PA-based sampling}: This method maintains the balance of privileged and unprivileged protected attributes within test data.
    \item {\bf WAE-based sampling}: This method maintains the balance of both class labels and protected attributes. Initially introduced in Fair-SMOTE\cite{Chakraborty2021BiasIM}, MAAT\cite{chen2022maat} also referred to it as the "We're All Equal" (WAE) method. 
\end{itemize}
Table \ref{tab:PSM-subgroup} presents the measurements of performance and fairness scores in the Adult dataset. In Table \ref{tab:subgroup2}, we provide the full set of comparisons conducted on each dataset. Using metric scores obtained on the original test data as the benchmark, we present the percentage of change in each metric. Further analysis of this experimental data is discussed later in \S\ref{result}.
% As shown in Table \ref{tab:PSM-subgroup}, the fairness scores are perfect among the samples matched via propensity scores. This indicates that the model is totally bias-free on the samples selected from PSM: The model fairly treats each matched pair, which consists of two data points with similar propensity scores and opposite protected attributes. On the other hand, fairness scores measured from the non-matched samples show dramatic bias, as shown in Table \ref{tab:PSM-subgroup}. 

% Based on the observations above, we argue that PSM is helpful in distinguishing test samples into two subgroups: (a) the subgroup where discrimination can barely be detected as test samples are equally treated regardless of the protected attributes and (b) the other group where samples with certain protected attributes are significantly discriminated, partially due to the fact the privileged and unprivileged groups do not share similar distributions in terms of class labels (as reflected by the propensity scores). 

\section{Proposed Approach}\label{proposed_approach}
In this section, we illustrate new methods that can be used to either measure or mitigate bias. 

\subsection{Controlled Fairness Testing}
As discussed in the previous section, we believe PSM provided insights regarding how different data points have different levels of risk of being exposed to discrimination. Therefore, when practitioners attempt to assess the extent of fairness of a model using testing samples, they should also consider the proportion of propensity matchable samples expected in future data. 

In light of providing more comprehensive insights from fairness testing, we propose to present fairness measures not as a singular score but as a breakdown projection. As shown in Fig \ref{fig:curve}, the curve demonstrates how the fairness scores vary as the proportion of propensity matchable samples change in testing data. For those who prefer a numerical score to represent the fairness level of the model, the projection curve can provide the fairness area under the curve (f-AUC) with similar intuition as the ROC AUC score. Given the specific fairness metric selected as the base metric, f-AUC can offer a comprehensive reflection on how the fairness scores might change along with the distribution drift in testing data.

We believe that controlled fairness testing is necessary, and the design brings value in many aspects:
\begin{itemize}
    \item {\bf Consistency}: Consistency ensures comparability. By controlling variables such as input data and test parameters, practitioners can accurately compare the fairness of different system outputs. If a testing set consists of an overwhelming amount of {\it unprivileged} group members, it is foreseeable that fairness metrics will be severely affected. However, such an imbalanced distribution may not be the expected speculation in the future deployment of the tested system.
    \item  {\bf Trade-off management}: By proactively identifying and addressing fairness concerns during the testing phase, organizations can reduce the likelihood of harmful outcomes and potential legal or reputational consequences. One way to achieve such moderation is to adjust the decision thresholds for different protected groups. However, such threshold calibration might be associated with a compromise of predictive performance. Therefore, controlled fairness testing plays an inevitably important role since it prevents practitioners from over-tuning the model and resulting in unnecessary and inferior trade-offs.
    \item {\bf Stakeholder accountability} Fairness testing is essential for building and maintaining trust among stakeholders, such as users, employees, and the general public. It is crucial to provide insights into the fairness level of complex systems in a transparent and accountable manner. Given most fairness metrics are data-driven, it is important to ensure stakeholders that fairness metrics are computed on testing data that is carefully composed. Moreover, stakeholders need to be informed of the cone of uncertainty: given the potential change in data distribution, it always remains possible that a model can be measured as more or less biased.
\end{itemize}
Overall, by systematically evaluating fairness under controlled conditions (estimated using PSM), practitioners can significantly enhance the accountability and trustworthiness of the tested ML software.

\begin{figure*}[!t]
\centering
\includegraphics[width=.9\textwidth]{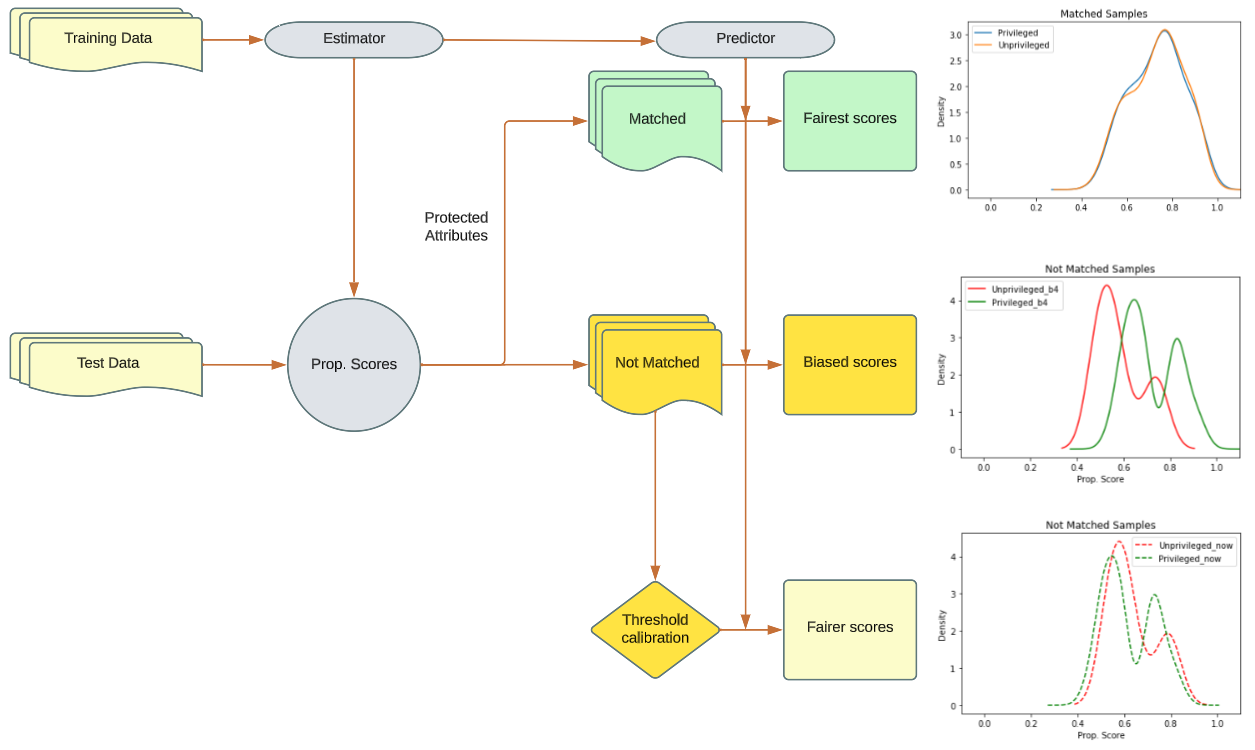}
% \caption{Fairness testing proposed in this paper. To evaluate the effectiveness of any de-biasing model, we want to ensure that the test set reflects the ideal data distribution in the future.}
\caption{Propensity score matching selects samples that represent the future distribution of class labels and the (expected) future distribution of protected attributes. }
\label{framework}\centering
\end{figure*}

\subsection{Selective Bias Mitigation}
As we discovered that one possible explanation for the biased model behavior could be the unmatched sample in the test set, we can propose a post-processing method leveraging the propensity scores to mitigate bias. In the new proposed methods, we use propensity scores to isolate samples that cannot be matched by propensity scores with regard to their protected attributes. We then calibrate the model's decision threshold (where the default is $I(P>0.5)=1$) such that the two distributions can become as indistinguishable as possible. As described in Figure \ref{framework} and Algorithm \ref{algo2}, we iteratively search for the optimal thresholds corresponding to each subgroup based on the protected attribute. The search generally follows two constraints: (1) The distribution of propensity scores after the threshold calibration should be indistinguishable as reflected by the statistical test. (2) While performing the calibration in constraint (1), the shifted thresholds for both groups should be as close to the original ones as possible. The formal formula Eq. \ref{eq1} and Eq. \ref{eq2} for the constraints are listed below:

\begin{equation}\label{eq1}
\begin{gathered}
\text{Maximize } p = P(|T| > |t'|) \\
\text{where }t' = \frac{(\bar{X}+\theta_{1}) - (\bar{Y}+\theta_{2})}{\sqrt{\frac{s_X^2}{n_X} + \frac{s_Y^2}{n_Y}}}
\end{gathered}
\end{equation}

\begin{equation}\label{eq2}
\begin{gathered}
\text{Minimize} |\theta_{1}|+|\theta_{2}|
\end{gathered}
\end{equation}
$X$ and $Y$ represent random variables from the privileged and unprivileged subgroups. $\theta_{1}$ and $\theta_{2}$ represent the calibrated thresholds applied to the corresponding subgroups. 

The intuition behind selective bias mitigation is that a biased model can still generate accurate and just outcomes for the majority of individuals, yet produce biased and inaccurate predictions for a subset of individuals. Therefore, when de-biasing the model, one needs to first identify data points with a higher likelihood of being discriminated. By leveraging PSM, we can now regard an isolated instance (that cannot be matched with a comparable instance with an opposite protected attribute) as a potential victim of model discrimination. As our proposed approach illustrates in Figure \ref{framework}, we can split the data waiting to be predicted into two subgroups:
\begin{itemize}
    \item {\bf Matched:} Pairs of data points that can be matched by propensity are those not suffering from model discrimination.
    \item {\bf Unmatched:} Instances that cannot find matchable opponents are more likely to be discriminated by the model. After computing the threshold gap between the two protected groups in the unmatched subset, we then adjust the decision thresholds for both the unprivileged and privileged groups.
\end{itemize}

\begin{algorithm}[t!]
\caption{Threshold Search \label{algo2}}
% \small
\KwData{$ps_{priv}, ps_{unpriv}$ contain propensity scores for the privileged and unprivileged groups respectively.
}
\KwResult{$\theta_{priv}, \theta_{unpriv}$ contain the final thresholds found by the iterative search guided by Eq. \ref{eq1} and Eq. \ref{eq2}.}
\Begin{
    $dist_{min} \gets$ $inf$ // minimum sum of $\theta_{priv}+\theta_{unpriv}$ \\ 
    $p_{max} \gets$ $0$ // largest p-value of t-test\\
    $pool_{priv} \gets [0:100]$\\
    $pool_{unpriv} \gets [0:100]$\\
    \While{$pool_{priv}.size \neq 0$}{
        $\theta_{1} \gets pool_{priv}$.pop()\\
            \While{$pool_{unpriv}.size \neq 0$}{
                $\theta_{2} \gets pool_{unpriv}$.pop()\\
                $ps_{priv}' \gets ps_{priv}$.add($\theta_{1}$)\\
                $ps_{unpriv}' \gets ps_{unpriv}$.subtract($\theta_{2}$)\\
                $pvalue \gets ttest(ps_{priv}',ps_{unpriv}')$\\
                \If{$pvalue > p_{max}$ and $abs(\theta_{1}+\theta_{2}) < dist_{min}$}{
                $p_{max} \gets pvalue$\\
                $dist_{min} \gets abs(\theta_{1}+\theta_{2})$\\
                $\theta_{priv}, \theta_{unpriv} \gets \theta_{1}, \theta_{2}$
                }
        }
    }
    return $\theta_{priv}, \theta_{unpriv}$
}
\end{algorithm}

\section{Experiment Setup}\label{experiment}

In this section, we describe the data preparation for the experiment as well as the general setup of the experiment.

In order to adhere to current empirical standards in SE,
that description adheres to the experimental design principles recommended in the ACM standards documents on empirical SE for optimization studies\footnote{\url{https://github.com/acmsigsoft/EmpiricalStandards/blob/master/docs/OptimizationStudies.md}}.

\subsection{Data}\label{data}
This paper uses data sets collected that are widely used in prior related research (see Table~\ref{tab:dataset}). After data collection, we first need to pre-process the data. For most of the datasets used in this paper (German, Bank, Heart, etc.), no feature engineering is required because either the features are all numerical or a standard procedure is adopted by all prior practitioners. As for others, in this experiment, a standardized pre-processing procedure is adopted, following guidelines from the AIF360 repository~\cite{bellamy2018ai}.
Finally, we apply min-max scaling (scale numerical values in the range of $[0,1]$ by the minimum and maximum values in each feature) to transform each data set. For each experiment trial, we divide the data into 70\% training data, and 30\% testing data, using the same set of random seeds in all methods to control the comparison variable. We repeat this procedure 20 times for statistical analysis. 

\begin{table}[t]
\small
\centering
\caption{Performance metrics based on binary confusion matrix.}
\begin{tabular}{cc}
\toprule
Metrics   & Definition                                    \\ \hline
Accuray   & (TP$+$TN)$/$(TP$+$TN$+$FP$+$FN)                         \\ 
Precision & TP$/$(TP$+$FP)                                    \\ 
Recall    & TP$/$(TP$+$FN)                                    \\ \
F1 score  & 2 $\times$ (Precision $\times$ Recall)$/$(Precision $+$ Recall) \\ 
% \hline
\bottomrule
\end{tabular}
\label{tab:metrics2}
\end{table}

\subsection{Evaluation Criteria}
To evaluate the predictive performance of each method, we use metrics computed by the binary classification confusion matrix: accuracy, precision, recall, and F1 score. These criteria are selected since they are widely used in both software analytics~\cite{local_tse,menzies2014sharing} and fairness literature~\cite{9286091,Biswas_2020, hardt2016equality, pleiss2017fairness, zhang2020white}. The definitions of the performance metrics are shown in Table~\ref{tab:metrics2}.
To assess the effectiveness of mitigating bias, we use the fairness metrics introduced in Table~\ref{tab:metrics}, some of which are also calculated based on the confusion matrix of binary classification. The group fairness metrics are designed to evaluate whether different social groups, as identified by their protected attributes, receive statistically similar prediction results by the classification model. 

Given that the performance and fairness metrics usually do not perfectly align with each other (e.g., model A might outperform model B in the metric DI, yet meanwhile lose in the metric AOD), Therefore, we use Generational Distance (GD) as an aggregated metric to reflect the overall quality of a model in terms of predictive performance as well as fairness. Widely used in multi-objective optimization \cite{peng2023veer}, GD computes the average distance, in terms of objective scores, between the solution set returned by a model and the actual optimal solution set. In this paper, we adopt and modify the equation of GD as follows:
\begin{equation}
    GD = \frac{1}{N}\sum_{i=1}^{N}\sqrt{(M_{i}-M_{i}^{*})^2}
\end{equation}
Where $N$ is the total number of performance/fairness metrics used in the evaluation, $M_{i}$ is the computed performance/fairness metric value, and $M_{i}^{*}$ is the optimal value for the corresponding metric (e.g., 100 for accuracy, 0 for AOD).

\begin{figure*}[!b]
        \centering
        \begin{subfigure}[b]{0.3\textwidth}
            \centering
            \includegraphics[width=\textwidth]{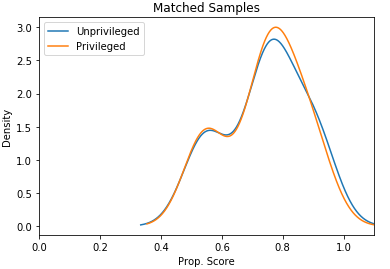}
            \caption[]%
            {{\small German dataset: matched  samples}}    
            \label{fig:german2}   \vspace{10mm}
        \end{subfigure}
        \hfill
        \begin{subfigure}[b]{0.3\textwidth}
            \centering
            \includegraphics[width=\textwidth]{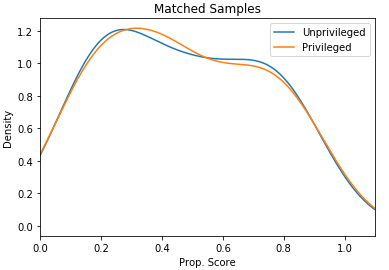}
            \caption[]%
            {{\small Heart dataset: matched  samples}}    
            \label{fig:heart2}   \vspace{10mm}
        \end{subfigure}
        \hfill
        \begin{subfigure}[b]{0.3\textwidth}  
            \centering 
            \includegraphics[width=\textwidth]{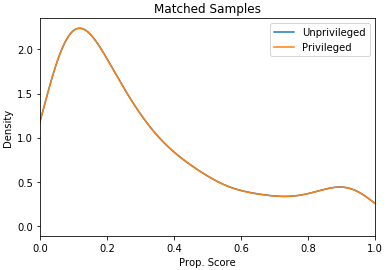}
            \caption[]%
            {{\small MEPS dataset: matched samples}}    
            \label{fig:meps2}  \vspace{10mm}
        \end{subfigure}
        \vspace{10mm}
        \begin{subfigure}[b]{0.3\textwidth}   
            \centering 
            \includegraphics[width=\textwidth]{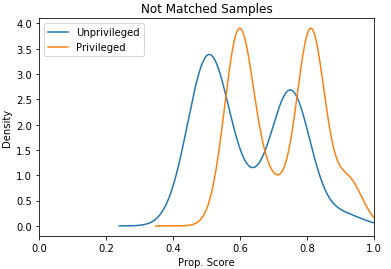}
            \caption[]%
            {{\small German dataset: unmatched samples}}    
            \label{fig:german1}
        \end{subfigure}
        \hfill
        \begin{subfigure}[b]{0.3\textwidth}   
            \centering 
            \includegraphics[width=\textwidth]{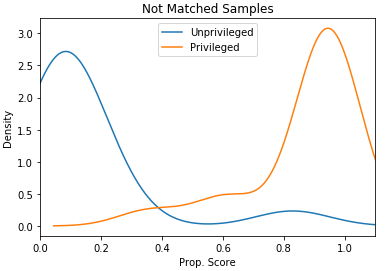}
            \caption[]%
            {{\small Heart dataset: unmatched samples}}    
            \label{fig:heart1}
        \end{subfigure}
        \hfill
        \begin{subfigure}[b]{0.3\textwidth}   
            \centering 
            \includegraphics[width=\textwidth]{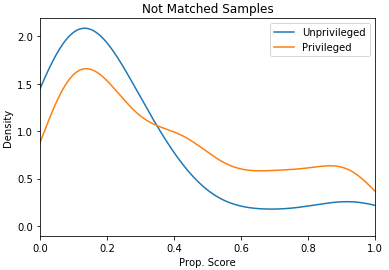}
            \caption[]%
            {{\small MEPS dataset: unmatched samples}}    
            \label{fig:meps1}
        \end{subfigure}
        \caption[ ]
        {\small Examples derived from three datasets. The distribution of propensity scores is drawn among the matched/unmatched samples.} 
        \label{fig:distribution}
    \end{figure*}

\subsection{Statistical Analysis}\label{stats}
To compare the predictive performance and ability to mitigate bias among all algorithms on every dataset, we use a non-parametric significance test along with a non-parametric effect size test. 
Specifically, we use the Scott-Knott test~\cite{mittas2012ranking} that sorts the list of treatments (in this case, the benchmark bias mitigation methods and our approach) by their median scores. After the sorting, it then splits the list into two sublists. The objective for such a split is to maximize the expected value of differences $E(\Delta)$ in the observed performances before and after division~\cite{xia2018hyperparameter}:
\begin{equation}
    E(\Delta) = \frac{|l_1|}{|l|}abs(E({l_1}) - E({l}))^2 + \frac{|l_2|}{|l|}abs(E({l_2}) - E({l}))^2
\end{equation}
where $|l_1|$ means the size of list $l_1$.

The Scott-Knott test assigns ranks to each result set; the higher the rank, the better the result. Two results will be ranked the same if the difference between the distributions is not significant.
In this expression, Cliff's Delta estimates the probability that a value in the list $A$ is greater than a value in the list $B$, minus the reverse probability~\cite{macbeth2011cliff}. A division passes this hypothesis test if it is not a ``small'' effect ($Delta \geq 0.147$). 
This hypothesis test and its effect sizes are supported by Hess and Kromery~\cite{hess2004robust}.

\subsection{Research Questions}
We assess the merits of {\IT} centering around the following research questions:

 {\bf RQ1:}  {\it Does propensity scores matter in fairness testing?}
Here, we explore rain-test set generation with/without reflecting on propensity
scores. Propensity will be shown to have a large impact on that data, so we will argue,
that we must not ignore propensity (since if we do, it introduces a threat to the validity of our conclusions).
 
    {\bf RQ2:} {\it To what extent is fairness testing influenced by different sampling strategies?} 
    % In this paper, we discuss whether the assumptions widely adopted in traditional testing schema (e.g., testing model performance on accuracy, recall, etc.) are applicable to fairness testing. In this RQ, we compare the fairness scores calculated on the original test set against those calculated on test samples selected by PSM.
    In this RQ, we wonder about the sensitivity of the fairness testing process when applying different sampling methods for selecting testing data. Each of the sampling methods is based on a corresponding assumption of future data: either the tendency of increasing balance in protected attributes or class labels, or both simultaneously. We also compare the magnitude of change in fairness scores against that in performance scores.
    
  {\bf RQ3:} {\it Is PSM a reliable testing mechanism? }
    The purpose of PSM is to reduce the confounding bias in observational studies. To achieve this, PSM examines whether the treatment is completely random, conditional on specified variables (in this case, protected attributes). Thus, we expect the samples selected by PSM to differ only from the original population in terms of protected attributes while the distribution of target classes remains the same.
    
    In this RQ, we ask if PSM-assisted testing can provide consistent and robust measures of the performance metrics. To answer this question, we compare the performance scores calculated on the original test set against those calculated on test samples selected by PSM. Our hypothesis is that if the testing set selected by PSM can constantly produce similar performance measures compared to those computed on the original testing test, we can claim that PSM is a reliable testing schema.
   
    % \item {\bf RQ3:} {\it How could PSM facilitate fairness testing?}
    % In this RQ, we examine a new fairness testing schema proposed in this paper. The new testing schema aims to offer more information to fairness testers concerning how fairness metrics can be affected as the future distribution of protected attributes changes.
    
   {\bf RQ4:} {\it Can we obtain superior trade-offs between performance and fairness?} We believe PSM can not only facilitate fairness testing but also help fix fairness bugs in ML software systems.
    In this RQ, we report the experimental results of a new bias mitigation method, FairMatch. This post-processing method adjusts the decision threshold in classification tasks for subgroups defined by their protected attributes. We analyze whether our proposed approach can outperform other bias mitigation methods.

\section{Results}\label{result}
  
\begin{blockquote}
\textbf{RQ1}: Is propensity score an important indicator in fairness testing?
\end{blockquote}
Our baseline research question is this: is it worthwhile exploring the issues
of this paper?  To answer this question, we discuss train-test set generation with/without reflecting on propensity scores. As we shall see, propensity leads to a large-scale
change in the nature of that train/test set-- which means that if researchers {\em ignore}
propensity, then their results are (a)~ influenced by something (propensity) that they are not
controlling in their experimental rig, and which (b)~can change and invalidate their
results. Hence, we argue that researchers should reflect on propensity lest they introduce a threat
to the validity of their results.

To make this point, we note that in observational studies, propensity scores matching refer to an alternative measurement when the random assignment of treatments to subjects is impossible in controlled experiments. Coincidentally, we found this design also fits in fairness testing: inevitably, we have to admit that it is impossible that the distributions of individuals from different protected groups are similarly random in the testing test.

This leads to an important question: does fairness testing use equally representative samples from different protected groups? And furthermore, how can we assess it? 
As shown in Fig. \ref{fig:distribution}, we compared the distribution of propensity scores between the unprivileged versus the privileged group. The right-hand-side figure shows a significant difference between the unprivileged versus privileged group among the non-matched samples. The privileged group in both sampled sets shares a rather similar distribution, whereas the two groups have completely different distributions in the non-matched samples. Thus, our answer to RQ1 is: {\bf Thanks to propensity scores, we have observed that a considerable part of testing data consisted of samples that are essentially incomparable. Therefore, it is reasonable to deduce that fairness testing relying on such testing data may not reflect the true quality of an ML model.}

\begin{table}[t!]
\centering
\small
\caption{RQ2 result: The average value and standard deviation of values in Table \ref{tab:subgroup2}.  }
\label{tab:rq2}
\resizebox{\columnwidth}{!}{%
\begin{tabular}{|l|cccc|cccc|}
\hline
AVG          & Acc   & Precision & Recall & F1    & aod   & eod   & spd   & di    \\ \hline
PSM\_sampled & 0.01  & 0.01      & 0.00   & 0.00  & 0.56  & -0.58 & -1.00 & -1.00 \\
C\_sampled   & 0.02  & -0.02     & -0.02  & -0.02 & 0.49  & 0.24  & -0.44 & -0.39 \\
PA\_sampled  & -0.11 & 0.06      & -0.01  & 0.00  & -0.30 & 0.14  & -0.21 & -0.32 \\
WAE\_sampled & -0.11 & 0.08      & -0.04  & -0.01 & 0.67  & 0.93  & -0.02 & -0.19 \\ \hline
STD          & Acc   & Precision & Recall & F1    & aod   & eod   & spd   & di    \\ \hline
PSM\_sampled & 0.05  & 0.05      & 0.12   & 0.06  & 2.54  & 0.46  & 0.00  & 0.00  \\
C\_sampled   & 0.05  & 0.17      & 0.07   & 0.11  & 1.33  & 0.57  & 0.34  & 0.35  \\
PA\_sampled  & 0.09  & 0.25      & 0.02   & 0.11  & 0.34  & 0.50  & 0.34  & 0.39  \\
WAE\_sampled & 0.10  & 0.23      & 0.10   & 0.11  & 1.42  & 1.51  & 0.43  & 0.21  \\ \hline
\end{tabular}
}
\end{table}

\begin{blockquote}
\textbf{RQ2}: To what extent is fairness testing influenced by different sampling strategies?
\end{blockquote}
As previously illustrated in Table \ref{tab:PSM-subgroup} and Table \ref{tab:subgroup2}, we evaluated the influence of four different sampling strategies on performance/fairness scores computed on the under-sampled test data. As illustrated in Table \ref{tab:subgroup2}, it is obvious that sampling methods tend to trade off performance for fairness in the majority cases. Furthermore, as presented in Table \ref{tab:rq2}, the average and standard deviation values show that the fluctuation in fairness scores is much more dramatic than that in performance scores. Such observation implies the necessity of exploring sampling methods in fairness testing: While selecting a subset of test data usually has a trivial influence on the final performance scores (in most cases, under 2\% change), the influence on fairness metrics is rather radical. It is noteworthy that the only case where the standard deviation is $0$ is when SPD and DI are constantly minimized by PSM sampling.

% Based on the observations above, we argue that PSM is helpful in distinguishing test samples into two subgroups: (a) the subgroup where discrimination can barely be detected as test samples are equally treated regardless of the protected attributes and (b) the other group where samples with certain protected attributes are significantly discriminated, partially due to the fact the privileged and unprivileged groups do not share similar distributions in terms of class labels (as reflected by the propensity scores).

In general, our answer to RQ2 is: {\bf Compared to performance metrics, fairness metrics show significantly greater variance in response to different sampling methods. Thus, we argue that fairness testing is more sensitive to sampling methods.}

\begin{figure}[!b]
        \centering
        \begin{subfigure}[b]{0.49\linewidth}
            \centering
            \includegraphics[width=\textwidth]{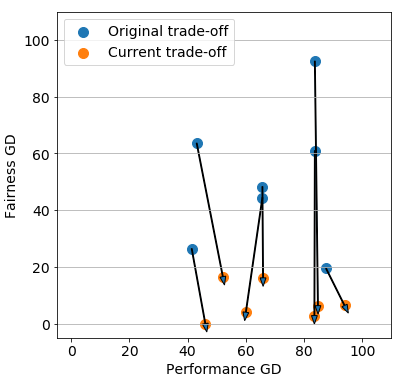}
            \caption[]%
            {{\small }}    
            \label{fig:pareto1}
        \end{subfigure}
        \hfill
        \begin{subfigure}[b]{0.49\linewidth}
            \centering
            \includegraphics[width=\textwidth]{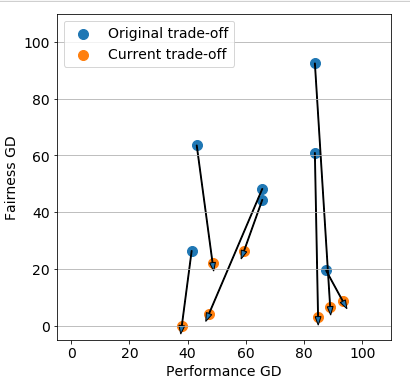}
            \caption[]%
            {{\small }}    
            \label{fig:pareto2}
        \end{subfigure}

        \caption[ ]
        {\small RQ3 result: The comparison of performance-fairness trade-offs before and after applying PSM. Each arrow represents a set of comparisons conducted on the datasets listed in Table \ref{tab:dataset}. For both axes, smaller values indicate better performance/fairness. } 
        \label{fig:pareto}
    \end{figure}
    
\begin{blockquote}
\textbf{RQ3}: Is PSM a reliable testing mechanism?
\end{blockquote}
In RQ1 and RQ2, we report that stratified sampling empowered by PSM significantly influences fairness scores of various kinds. Naturally, we wonder if a similar influence will be observed on the performance scores. Therefore, we calculated the changes in both fairness and performance scores with respect to the change in PSM samples. Each generational distance (GD)is computed based on 4 fairness/performance metrics, respectively: AOD, EOD, SPD, and DI are used to calculate fairness GDs; accuracy, precision, recall, and f1 scores are used to calculate performance GDs. The GD represents the Euclidean distance between the current fairness-performance trade-off achieved by the model and the optimal fairness-performance trade-off (e.g., AOD=0 and Accuracy=100). It is noteworthy that the optimal trade-off is usually impossible to obtain, but used as a reference point to calculate GDs.

As shown in Fig \ref{fig:pareto}, we show how the reported fairness-performance trade-offs change. We found that, compared to the dramatic changes in fairness GDs, changes in performance GDs are rather trivial. In some cases, the fairness GD experiences a $90\%$ drop while the change in performance GD is below $1\%$. This indicates that our proposed testing schema is robust in reporting the predictive performance of a model. Our answer to RQ3 is: {\bf While trivial fluctuations are observed, the new testing schema will not affect performance testing results.}

\begin{table*}[]
\caption{Results for RQ4.  For all performance metrics, greater is better; for all fairness metrics, smaller is better. Here, cells marked in darker colors are better than those marked in lighter colors within the same dataset block. The ranks indicated by colors are determined by the Scott-Knott test. "LR" denotes the baseline logistic regression model.}
\label{rq3-1}
\resizebox{\textwidth}{!}{%
\begin{tabular}{|c|cccccc|cccccc|cccccc|cccccc|}
\hline
Dataset &
  \multicolumn{6}{c|}{Adult: Sex} &
  \multicolumn{6}{c|}{Adult: Race} &
  \multicolumn{6}{c|}{Compas: Sex} &
  \multicolumn{6}{c|}{Compas: Race} \\ \hline
Method &
  LR &
  EGR &
  FAXAI &
  FairMask &
  MAAT &
  \textbf{FairMatch} &
  LR &
  EGR &
  FAXAI &
  FairMask &
  MAAT &
  \textbf{FairMatch} &
  LR &
  EGR &
  FAXAI &
  FairMask &
  MAAT &
  \textbf{FairMatch} &
  LR &
  EGR &
  FAXAI &
  FairMask &
  MAAT &
  \textbf{FairMatch} \\ \hline
Accuracy &
  \cellcolor[HTML]{969696}84 &
  \cellcolor[HTML]{969696}84 &
  \cellcolor[HTML]{969696}84 &
  \cellcolor[HTML]{969696}84 &
  \cellcolor[HTML]{969696}84 &
  \cellcolor[HTML]{969696}84 &
  \cellcolor[HTML]{969696}85 &
  \cellcolor[HTML]{969696}84 &
  \cellcolor[HTML]{969696}85 &
  \cellcolor[HTML]{969696}85 &
  \cellcolor[HTML]{969696}84 &
  \cellcolor[HTML]{969696}85 &
  \cellcolor[HTML]{969696}66 &
  \cellcolor[HTML]{969696}65 &
  \cellcolor[HTML]{969696}66 &
  \cellcolor[HTML]{969696}66 &
  \cellcolor[HTML]{969696}67 &
  \cellcolor[HTML]{969696}65 &
  \cellcolor[HTML]{969696}66 &
  \cellcolor[HTML]{969696}66 &
  \cellcolor[HTML]{969696}66 &
  \cellcolor[HTML]{969696}66 &
  \cellcolor[HTML]{969696}67 &
  \cellcolor[HTML]{969696}65 \\
Precision &
  \cellcolor[HTML]{969696}73 &
  \cellcolor[HTML]{969696}73 &
  \cellcolor[HTML]{969696}73 &
  \cellcolor[HTML]{969696}73 &
  \cellcolor[HTML]{969696}72 &
  \cellcolor[HTML]{BBBBBB}65 &
  \cellcolor[HTML]{969696}74 &
  \cellcolor[HTML]{969696}72 &
  \cellcolor[HTML]{969696}73 &
  \cellcolor[HTML]{969696}73 &
  \cellcolor[HTML]{969696}72 &
  \cellcolor[HTML]{969696}73 &
  \cellcolor[HTML]{969696}67 &
  \cellcolor[HTML]{969696}66 &
  \cellcolor[HTML]{969696}66 &
  \cellcolor[HTML]{969696}67 &
  \cellcolor[HTML]{969696}65 &
  \cellcolor[HTML]{BBBBBB}64 &
  \cellcolor[HTML]{969696}67 &
  \cellcolor[HTML]{969696}65 &
  \cellcolor[HTML]{969696}67 &
  \cellcolor[HTML]{969696}67 &
  \cellcolor[HTML]{969696}66 &
  \cellcolor[HTML]{969696}65 \\
Recall &
  \cellcolor[HTML]{FFFFFF}59 &
  \cellcolor[HTML]{FFFFFF}54 &
  \cellcolor[HTML]{FFFFFF}56 &
  \cellcolor[HTML]{FFFFFF}59 &
  \cellcolor[HTML]{969696}81 &
  \cellcolor[HTML]{BBBBBB}73 &
  \cellcolor[HTML]{BBBBBB}59 &
  \cellcolor[HTML]{BBBBBB}59 &
  \cellcolor[HTML]{BBBBBB}60 &
  \cellcolor[HTML]{BBBBBB}60 &
  \cellcolor[HTML]{969696}81 &
  \cellcolor[HTML]{BBBBBB}60 &
  \cellcolor[HTML]{BBBBBB}74 &
  \cellcolor[HTML]{BBBBBB}76 &
  \cellcolor[HTML]{BBBBBB}76 &
  \cellcolor[HTML]{BBBBBB}75 &
  \cellcolor[HTML]{FFFFFF}67 &
  \cellcolor[HTML]{969696}82 &
  \cellcolor[HTML]{BBBBBB}74 &
  \cellcolor[HTML]{BBBBBB}78 &
  \cellcolor[HTML]{BBBBBB}76 &
  \cellcolor[HTML]{BBBBBB}75 &
  \cellcolor[HTML]{FFFFFF}67 &
  \cellcolor[HTML]{969696}80 \\
F1 &
  \cellcolor[HTML]{BBBBBB}66 &
  \cellcolor[HTML]{BBBBBB}62 &
  \cellcolor[HTML]{BBBBBB}64 &
  \cellcolor[HTML]{BBBBBB}66 &
  \cellcolor[HTML]{969696}75 &
  \cellcolor[HTML]{BBBBBB}68 &
  \cellcolor[HTML]{BBBBBB}66 &
  \cellcolor[HTML]{BBBBBB}65 &
  \cellcolor[HTML]{BBBBBB}66 &
  \cellcolor[HTML]{BBBBBB}66 &
  \cellcolor[HTML]{969696}75 &
  \cellcolor[HTML]{BBBBBB}66 &
  \cellcolor[HTML]{969696}70 &
  \cellcolor[HTML]{969696}70 &
  \cellcolor[HTML]{969696}71 &
  \cellcolor[HTML]{969696}71 &
  \cellcolor[HTML]{BBBBBB}65 &
  \cellcolor[HTML]{969696}72 &
  \cellcolor[HTML]{969696}70 &
  \cellcolor[HTML]{969696}70 &
  \cellcolor[HTML]{969696}71 &
  \cellcolor[HTML]{969696}71 &
  \cellcolor[HTML]{BBBBBB}65 &
  \cellcolor[HTML]{969696}72 \\
AOD &
  \cellcolor[HTML]{969696}2 &
  \cellcolor[HTML]{969696}2 &
  \cellcolor[HTML]{969696}3 &
  \cellcolor[HTML]{969696}2 &
  \cellcolor[HTML]{BBBBBB}4 &
  \cellcolor[HTML]{969696}2 &
  4 &
  \cellcolor[HTML]{BBBBBB}2 &
  \cellcolor[HTML]{BBBBBB}2 &
  \cellcolor[HTML]{BBBBBB}2 &
  \cellcolor[HTML]{BBBBBB}2 &
  \cellcolor[HTML]{969696}1 &
  \cellcolor[HTML]{BBBBBB}6 &
  \cellcolor[HTML]{BBBBBB}4 &
  \cellcolor[HTML]{969696}2 &
  \cellcolor[HTML]{BBBBBB}4 &
  \cellcolor[HTML]{FFFFFF}12 &
  \cellcolor[HTML]{BBBBBB}5 &
  \cellcolor[HTML]{BBBBBB}6 &
  \cellcolor[HTML]{969696}3 &
  \cellcolor[HTML]{BBBBBB}6 &
  \cellcolor[HTML]{969696}3 &
  \cellcolor[HTML]{BBBBBB}7 &
  \cellcolor[HTML]{969696}4 \\
EOD &
  12 &
  \cellcolor[HTML]{969696}2 &
  \cellcolor[HTML]{969696}4 &
  \cellcolor[HTML]{BBBBBB}10 &
  \cellcolor[HTML]{969696}4 &
  \cellcolor[HTML]{BBBBBB}9 &
  13 &
  \cellcolor[HTML]{969696}3 &
  \cellcolor[HTML]{969696}6 &
  \cellcolor[HTML]{BBBBBB}6 &
  \cellcolor[HTML]{969696}4 &
  \cellcolor[HTML]{969696}4 &
  24 &
  \cellcolor[HTML]{969696}4 &
  \cellcolor[HTML]{BBBBBB}10 &
  \cellcolor[HTML]{969696}6 &
  \cellcolor[HTML]{969696}7 &
  \cellcolor[HTML]{969696}6 &
  11 &
  \cellcolor[HTML]{969696}4 &
  \cellcolor[HTML]{BBBBBB}9 &
  \cellcolor[HTML]{BBBBBB}6 &
  \cellcolor[HTML]{969696}3 &
  \cellcolor[HTML]{969696}4 \\
SPD &
  \cellcolor[HTML]{BBBBBB}19 &
  \cellcolor[HTML]{969696}12 &
  \cellcolor[HTML]{969696}13 &
  \cellcolor[HTML]{BBBBBB}18 &
  \cellcolor[HTML]{969696}11 &
  \cellcolor[HTML]{969696}14 &
  11 &
  \cellcolor[HTML]{969696}7 &
  \cellcolor[HTML]{969696}9 &
  \cellcolor[HTML]{BBBBBB}9 &
  \cellcolor[HTML]{969696}7 &
  \cellcolor[HTML]{969696}7 &
  33 &
  \cellcolor[HTML]{969696}4 &
  \cellcolor[HTML]{BBBBBB}15 &
  \cellcolor[HTML]{BBBBBB}12 &
  \cellcolor[HTML]{BBBBBB}16 &
  \cellcolor[HTML]{BBBBBB}14 &
  20 &
  \cellcolor[HTML]{BBBBBB}8 &
  \cellcolor[HTML]{FFFFFF}16 &
  \cellcolor[HTML]{FFFFFF}15 &
  \cellcolor[HTML]{BBBBBB}9 &
  \cellcolor[HTML]{969696}3 \\
DI &
  74 &
  \cellcolor[HTML]{969696}50 &
  \cellcolor[HTML]{BBBBBB}57 &
  \cellcolor[HTML]{BBBBBB}68 &
  \cellcolor[HTML]{BBBBBB}57 &
  \cellcolor[HTML]{969696}52 &
  52 &
  \cellcolor[HTML]{969696}33 &
  \cellcolor[HTML]{BBBBBB}41 &
  \cellcolor[HTML]{BBBBBB}44 &
  \cellcolor[HTML]{969696}33 &
  \cellcolor[HTML]{969696}28 &
  37 &
  \cellcolor[HTML]{969696}6 &
  \cellcolor[HTML]{BBBBBB}20 &
  \cellcolor[HTML]{BBBBBB}18 &
  \cellcolor[HTML]{BBBBBB}19 &
  \cellcolor[HTML]{BBBBBB}17 &
  27 &
  \cellcolor[HTML]{BBBBBB}11 &
  \cellcolor[HTML]{FFFFFF}23 &
  \cellcolor[HTML]{FFFFFF}21 &
  \cellcolor[HTML]{BBBBBB}13 &
  \cellcolor[HTML]{969696}5 \\ \hline
Dataset &
  \multicolumn{6}{c|}{Bank: Age} &
  \multicolumn{6}{c|}{German: Sex} &
  \multicolumn{6}{c|}{Health: Age} &
  \multicolumn{6}{c|}{MEPS: Race} \\ \hline
Method &
  LR &
  EGR &
  FAXAI &
  FairMask &
  MAAT &
  \textbf{FairMatch} &
  LR &
  EGR &
  FAXAI &
  FairMask &
  MAAT &
  \textbf{FairMatch} &
  LR &
  EGR &
  FAXAI &
  FairMask &
  MAAT &
  \textbf{FairMatch} &
  LR &
  EGR &
  FAXAI &
  FairMask &
  MAAT &
  \textbf{FairMatch} \\ \hline
Accuracy &
  \cellcolor[HTML]{969696}90 &
  \cellcolor[HTML]{969696}90 &
  \cellcolor[HTML]{969696}90 &
  \cellcolor[HTML]{969696}90 &
  \cellcolor[HTML]{969696}89 &
  \cellcolor[HTML]{969696}85 &
  \cellcolor[HTML]{969696}77 &
  \cellcolor[HTML]{969696}75 &
  \cellcolor[HTML]{969696}75 &
  \cellcolor[HTML]{969696}76 &
  \cellcolor[HTML]{969696}75 &
  \cellcolor[HTML]{969696}75 &
  \cellcolor[HTML]{969696}83 &
  \cellcolor[HTML]{969696}83 &
  \cellcolor[HTML]{969696}83 &
  \cellcolor[HTML]{969696}83 &
  \cellcolor[HTML]{969696}83 &
  \cellcolor[HTML]{969696}82 &
  \cellcolor[HTML]{969696}87 &
  \cellcolor[HTML]{969696}86 &
  \cellcolor[HTML]{969696}86 &
  \cellcolor[HTML]{969696}87 &
  \cellcolor[HTML]{969696}87 &
  \cellcolor[HTML]{969696}88 \\
Precision &
  \cellcolor[HTML]{969696}67 &
  \cellcolor[HTML]{969696}66 &
  \cellcolor[HTML]{969696}67 &
  \cellcolor[HTML]{969696}67 &
  \cellcolor[HTML]{969696}68 &
  \cellcolor[HTML]{BBBBBB}46 &
  \cellcolor[HTML]{BBBBBB}80 &
  \cellcolor[HTML]{BBBBBB}79 &
  \cellcolor[HTML]{BBBBBB}78 &
  \cellcolor[HTML]{BBBBBB}80 &
  \cellcolor[HTML]{FFFFFF}66 &
  \cellcolor[HTML]{969696}83 &
  \cellcolor[HTML]{969696}80 &
  \cellcolor[HTML]{969696}80 &
  \cellcolor[HTML]{969696}80 &
  \cellcolor[HTML]{969696}82 &
  \cellcolor[HTML]{969696}82 &
  \cellcolor[HTML]{969696}82 &
  \cellcolor[HTML]{969696}71 &
  \cellcolor[HTML]{BBBBBB}67 &
  \cellcolor[HTML]{969696}70 &
  \cellcolor[HTML]{969696}71 &
  \cellcolor[HTML]{BBBBBB}67 &
  53 \\
Recall &
  \cellcolor[HTML]{FFFFFF}40 &
  \cellcolor[HTML]{FFFFFF}40 &
  \cellcolor[HTML]{FFFFFF}39 &
  \cellcolor[HTML]{FFFFFF}40 &
  \cellcolor[HTML]{BBBBBB}76 &
  \cellcolor[HTML]{969696}87 &
  \cellcolor[HTML]{969696}88 &
  \cellcolor[HTML]{969696}87 &
  \cellcolor[HTML]{969696}89 &
  \cellcolor[HTML]{969696}87 &
  \cellcolor[HTML]{FFFFFF}70 &
  \cellcolor[HTML]{BBBBBB}78 &
  \cellcolor[HTML]{969696}84 &
  \cellcolor[HTML]{969696}81 &
  \cellcolor[HTML]{969696}81 &
  \cellcolor[HTML]{969696}83 &
  \cellcolor[HTML]{969696}82 &
  \cellcolor[HTML]{969696}80 &
  \cellcolor[HTML]{FFFFFF}42 &
  35 &
  38 &
  \cellcolor[HTML]{FFFFFF}40 &
  \cellcolor[HTML]{BBBBBB}79 &
  \cellcolor[HTML]{BBBBBB}53 \\
F1 &
  \cellcolor[HTML]{FFFFFF}50 &
  \cellcolor[HTML]{FFFFFF}50 &
  \cellcolor[HTML]{FFFFFF}50 &
  \cellcolor[HTML]{FFFFFF}50 &
  \cellcolor[HTML]{969696}72 &
  \cellcolor[HTML]{BBBBBB}60 &
  \cellcolor[HTML]{969696}83 &
  \cellcolor[HTML]{969696}82 &
  \cellcolor[HTML]{969696}83 &
  \cellcolor[HTML]{969696}83 &
  \cellcolor[HTML]{BBBBBB}68 &
  \cellcolor[HTML]{969696}81 &
  \cellcolor[HTML]{969696}82 &
  \cellcolor[HTML]{969696}81 &
  \cellcolor[HTML]{969696}81 &
  \cellcolor[HTML]{969696}82 &
  \cellcolor[HTML]{969696}82 &
  \cellcolor[HTML]{969696}81 &
  \cellcolor[HTML]{BBBBBB}53 &
  \cellcolor[HTML]{FFFFFF}46 &
  \cellcolor[HTML]{FFFFFF}49 &
  \cellcolor[HTML]{BBBBBB}51 &
  \cellcolor[HTML]{969696}70 &
  \cellcolor[HTML]{BBBBBB}53 \\
AOD &
  6 &
  \cellcolor[HTML]{969696}2 &
  \cellcolor[HTML]{969696}3 &
  \cellcolor[HTML]{BBBBBB}5 &
  \cellcolor[HTML]{BBBBBB}4 &
  \cellcolor[HTML]{BBBBBB}4 &
  \cellcolor[HTML]{BBBBBB}7 &
  \cellcolor[HTML]{969696}4 &
  \cellcolor[HTML]{969696}4 &
  \cellcolor[HTML]{969696}4 &
  \cellcolor[HTML]{969696}5 &
  \cellcolor[HTML]{969696}4 &
  11 &
  10 &
  11 &
  \cellcolor[HTML]{BBBBBB}7 &
  \cellcolor[HTML]{FFFFFF}11 &
  \cellcolor[HTML]{969696}3 &
  \cellcolor[HTML]{BBBBBB}8 &
  \cellcolor[HTML]{969696}3 &
  \cellcolor[HTML]{969696}6 &
  \cellcolor[HTML]{969696}4 &
  \cellcolor[HTML]{969696}7 &
  \cellcolor[HTML]{969696}4 \\
EOD &
  17 &
  \cellcolor[HTML]{969696}5 &
  \cellcolor[HTML]{969696}7 &
  \cellcolor[HTML]{BBBBBB}14 &
  \cellcolor[HTML]{969696}5 &
  \cellcolor[HTML]{969696}6 &
  \cellcolor[HTML]{BBBBBB}6 &
  \cellcolor[HTML]{969696}3 &
  \cellcolor[HTML]{BBBBBB}5 &
  \cellcolor[HTML]{BBBBBB}5 &
  \cellcolor[HTML]{BBBBBB}6 &
  \cellcolor[HTML]{969696}3 &
  \cellcolor[HTML]{BBBBBB}12 &
  \cellcolor[HTML]{BBBBBB}10 &
  \cellcolor[HTML]{BBBBBB}10 &
  \cellcolor[HTML]{BBBBBB}10 &
  \cellcolor[HTML]{BBBBBB}11 &
  \cellcolor[HTML]{969696}6 &
  23 &
  \cellcolor[HTML]{969696}5 &
  \cellcolor[HTML]{BBBBBB}15 &
  \cellcolor[HTML]{BBBBBB}10 &
  \cellcolor[HTML]{BBBBBB}12 &
  \cellcolor[HTML]{969696}4 \\
SPD &
  11 &
  \cellcolor[HTML]{969696}5 &
  \cellcolor[HTML]{969696}7 &
  \cellcolor[HTML]{BBBBBB}9 &
  \cellcolor[HTML]{969696}5 &
  \cellcolor[HTML]{FFFFFF}11 &
  \cellcolor[HTML]{BBBBBB}12 &
  \cellcolor[HTML]{969696}5 &
  \cellcolor[HTML]{969696}4 &
  \cellcolor[HTML]{969696}5 &
  \cellcolor[HTML]{969696}7 &
  \cellcolor[HTML]{969696}4 &
  41 &
  \cellcolor[HTML]{969696}24 &
  \cellcolor[HTML]{969696}23 &
  \cellcolor[HTML]{969696}23 &
  \cellcolor[HTML]{BBBBBB}28 &
  \cellcolor[HTML]{969696}24 &
  14 &
  \cellcolor[HTML]{969696}5 &
  \cellcolor[HTML]{BBBBBB}9 &
  \cellcolor[HTML]{BBBBBB}9 &
  \cellcolor[HTML]{BBBBBB}9 &
  \cellcolor[HTML]{969696}4 \\
DI &
  151 &
  \cellcolor[HTML]{969696}68 &
  \cellcolor[HTML]{969696}91 &
  \cellcolor[HTML]{BBBBBB}130 &
  \cellcolor[HTML]{FFFFFF}156 &
  \cellcolor[HTML]{969696}91 &
  \cellcolor[HTML]{BBBBBB}15 &
  \cellcolor[HTML]{969696}6 &
  \cellcolor[HTML]{969696}6 &
  \cellcolor[HTML]{969696}6 &
  \cellcolor[HTML]{BBBBBB}11 &
  \cellcolor[HTML]{969696}6 &
  60 &
  \cellcolor[HTML]{969696}43 &
  \cellcolor[HTML]{969696}43 &
  \cellcolor[HTML]{BBBBBB}50 &
  \cellcolor[HTML]{BBBBBB}51 &
  \cellcolor[HTML]{969696}44 &
  68 &
  \cellcolor[HTML]{BBBBBB}37 &
  62 &
  57 &
  59 &
  \cellcolor[HTML]{969696}20 \\ \hline
\end{tabular}%
}
\end{table*}

\begin{table*}[]
\caption{More Results for RQ4.  For all performance metrics, greater is better; for all fairness metrics, smaller is better. Here, cells marked in darker colors are better than those marked in lighter colors within the same dataset block. "XGB" denotes the baseline gradient boosting model.}
\label{rq3-2}
\resizebox{\textwidth}{!}{%
\begin{tabular}{|c|cccccc|cccccc|cccccc|cccccc|}
\hline
Dataset &
  \multicolumn{6}{c|}{Adult: Sex} &
  \multicolumn{6}{c|}{Adult: Race} &
  \multicolumn{6}{c|}{Compas: Sex} &
  \multicolumn{6}{c|}{Compas: Race} \\ \hline
Method &
  XGB &
  EGR &
  FAXAI &
  FairMask &
  MAAT &
  \textbf{FairMatch} &
  XGB &
  EGR &
  FAXAI &
  FairMask &
  MAAT &
  \textbf{FairMatch} &
  XGB &
  EGR &
  FAXAI &
  FairMask &
  MAAT &
  \textbf{FairMatch} &
  XGB &
  EGR &
  FAXAI &
  FairMask &
  MAAT &
  \textbf{FairMatch} \\ \hline
Accuracy &
  \cellcolor[HTML]{969696}86 &
  \cellcolor[HTML]{969696}85 &
  \cellcolor[HTML]{969696}86 &
  \cellcolor[HTML]{969696}86 &
  \cellcolor[HTML]{969696}86 &
  \cellcolor[HTML]{969696}86 &
  \cellcolor[HTML]{969696}86 &
  \cellcolor[HTML]{969696}85 &
  \cellcolor[HTML]{969696}86 &
  \cellcolor[HTML]{969696}86 &
  \cellcolor[HTML]{969696}86 &
  \cellcolor[HTML]{969696}86 &
  \cellcolor[HTML]{969696}68 &
  \cellcolor[HTML]{969696}66 &
  \cellcolor[HTML]{969696}68 &
  \cellcolor[HTML]{969696}68 &
  \cellcolor[HTML]{969696}69 &
  \cellcolor[HTML]{969696}67 &
  \cellcolor[HTML]{969696}68 &
  \cellcolor[HTML]{969696}66 &
  \cellcolor[HTML]{969696}68 &
  \cellcolor[HTML]{969696}68 &
  \cellcolor[HTML]{969696}68 &
  \cellcolor[HTML]{969696}68 \\
Precision &
  \cellcolor[HTML]{969696}78 &
  \cellcolor[HTML]{969696}77 &
  \cellcolor[HTML]{969696}79 &
  \cellcolor[HTML]{969696}79 &
  \cellcolor[HTML]{BBBBBB}74 &
  \cellcolor[HTML]{969696}77 &
  \cellcolor[HTML]{969696}78 &
  \cellcolor[HTML]{BBBBBB}73 &
  \cellcolor[HTML]{969696}79 &
  \cellcolor[HTML]{969696}78 &
  \cellcolor[HTML]{969696}76 &
  \cellcolor[HTML]{969696}79 &
  \cellcolor[HTML]{969696}67 &
  \cellcolor[HTML]{969696}67 &
  \cellcolor[HTML]{969696}69 &
  \cellcolor[HTML]{969696}68 &
  \cellcolor[HTML]{969696}68 &
  \cellcolor[HTML]{969696}68 &
  \cellcolor[HTML]{969696}67 &
  \cellcolor[HTML]{969696}67 &
  \cellcolor[HTML]{969696}69 &
  \cellcolor[HTML]{969696}68 &
  \cellcolor[HTML]{969696}68 &
  \cellcolor[HTML]{969696}69 \\
Recall &
  \cellcolor[HTML]{BBBBBB}60 &
  \cellcolor[HTML]{BBBBBB}57 &
  \cellcolor[HTML]{BBBBBB}59 &
  \cellcolor[HTML]{BBBBBB}61 &
  \cellcolor[HTML]{969696}85 &
  \cellcolor[HTML]{BBBBBB}61 &
  \cellcolor[HTML]{BBBBBB}60 &
  \cellcolor[HTML]{BBBBBB}60 &
  \cellcolor[HTML]{BBBBBB}60 &
  \cellcolor[HTML]{BBBBBB}60 &
  \cellcolor[HTML]{969696}84 &
  \cellcolor[HTML]{BBBBBB}61 &
  \cellcolor[HTML]{969696}77 &
  \cellcolor[HTML]{969696}76 &
  \cellcolor[HTML]{969696}75 &
  \cellcolor[HTML]{969696}75 &
  \cellcolor[HTML]{BBBBBB}69 &
  \cellcolor[HTML]{969696}74 &
  \cellcolor[HTML]{969696}77 &
  \cellcolor[HTML]{969696}76 &
  \cellcolor[HTML]{969696}75 &
  \cellcolor[HTML]{969696}76 &
  \cellcolor[HTML]{BBBBBB}68 &
  \cellcolor[HTML]{969696}76 \\
F1 &
  \cellcolor[HTML]{BBBBBB}68 &
  \cellcolor[HTML]{BBBBBB}66 &
  \cellcolor[HTML]{BBBBBB}68 &
  \cellcolor[HTML]{BBBBBB}69 &
  \cellcolor[HTML]{969696}79 &
  \cellcolor[HTML]{BBBBBB}68 &
  \cellcolor[HTML]{BBBBBB}68 &
  \cellcolor[HTML]{BBBBBB}65 &
  \cellcolor[HTML]{BBBBBB}68 &
  \cellcolor[HTML]{BBBBBB}67 &
  \cellcolor[HTML]{969696}79 &
  \cellcolor[HTML]{BBBBBB}69 &
  \cellcolor[HTML]{969696}72 &
  \cellcolor[HTML]{969696}71 &
  \cellcolor[HTML]{969696}72 &
  \cellcolor[HTML]{969696}72 &
  \cellcolor[HTML]{BBBBBB}68 &
  \cellcolor[HTML]{969696}71 &
  \cellcolor[HTML]{969696}72 &
  \cellcolor[HTML]{969696}71 &
  \cellcolor[HTML]{969696}72 &
  \cellcolor[HTML]{969696}72 &
  \cellcolor[HTML]{BBBBBB}68 &
  \cellcolor[HTML]{969696}72 \\
AOD &
  \cellcolor[HTML]{BBBBBB}4 &
  \cellcolor[HTML]{969696}2 &
  \cellcolor[HTML]{969696}1 &
  \cellcolor[HTML]{969696}2 &
  \cellcolor[HTML]{969696}4 &
  \cellcolor[HTML]{969696}1 &
  \cellcolor[HTML]{BBBBBB}2 &
  \cellcolor[HTML]{969696}1 &
  \cellcolor[HTML]{BBBBBB}2 &
  \cellcolor[HTML]{969696}1 &
  \cellcolor[HTML]{BBBBBB}2 &
  \cellcolor[HTML]{969696}1 &
  \cellcolor[HTML]{BBBBBB}4 &
  \cellcolor[HTML]{969696}2 &
  \cellcolor[HTML]{969696}3 &
  \cellcolor[HTML]{969696}2 &
  \cellcolor[HTML]{FFFFFF}14 &
  \cellcolor[HTML]{969696}3 &
  \cellcolor[HTML]{969696}3 &
  \cellcolor[HTML]{BBBBBB}4 &
  \cellcolor[HTML]{BBBBBB}4 &
  \cellcolor[HTML]{969696}2 &
  \cellcolor[HTML]{FFFFFF}13 &
  \cellcolor[HTML]{969696}3 \\
EOD &
  10 &
  \cellcolor[HTML]{969696}2 &
  8 &
  8 &
  3 &
  \cellcolor[HTML]{BBBBBB}6 &
  5 &
  \cellcolor[HTML]{BBBBBB}3 &
  \cellcolor[HTML]{BBBBBB}4 &
  \cellcolor[HTML]{969696}2 &
  \cellcolor[HTML]{969696}2 &
  \cellcolor[HTML]{969696}2 &
  10 &
  \cellcolor[HTML]{969696}4 &
  \cellcolor[HTML]{BBBBBB}8 &
  \cellcolor[HTML]{BBBBBB}9 &
  \cellcolor[HTML]{FFFFFF}12 &
  \cellcolor[HTML]{969696}5 &
  12 &
  \cellcolor[HTML]{969696}3 &
  11 &
  \cellcolor[HTML]{BBBBBB}10 &
  \cellcolor[HTML]{BBBBBB}8 &
  \cellcolor[HTML]{969696}4 \\
SPD &
  17 &
  \cellcolor[HTML]{969696}12 &
  16 &
  17 &
  12 &
  \cellcolor[HTML]{969696}14 &
  \cellcolor[HTML]{BBBBBB}8 &
  \cellcolor[HTML]{969696}6 &
  \cellcolor[HTML]{BBBBBB}8 &
  \cellcolor[HTML]{BBBBBB}8 &
  \cellcolor[HTML]{BBBBBB}8 &
  \cellcolor[HTML]{969696}7 &
  17 &
  \cellcolor[HTML]{969696}8 &
  \cellcolor[HTML]{BBBBBB}15 &
  \cellcolor[HTML]{BBBBBB}14 &
  \cellcolor[HTML]{FFFFFF}18 &
  \cellcolor[HTML]{969696}9 &
  18 &
  \cellcolor[HTML]{969696}4 &
  17 &
  \cellcolor[HTML]{BBBBBB}14 &
  \cellcolor[HTML]{FFFFFF}15 &
  \cellcolor[HTML]{969696}9 \\
DI &
  72 &
  \cellcolor[HTML]{969696}50 &
  69 &
  65 &
  50 &
  \cellcolor[HTML]{BBBBBB}60 &
  40 &
  \cellcolor[HTML]{969696}31 &
  \cellcolor[HTML]{BBBBBB}37 &
  39 &
  39 &
  \cellcolor[HTML]{BBBBBB}37 &
  25 &
  \cellcolor[HTML]{969696}11 &
  \cellcolor[HTML]{BBBBBB}21 &
  \cellcolor[HTML]{BBBBBB}21 &
  \cellcolor[HTML]{FFFFFF}23 &
  \cellcolor[HTML]{969696}13 &
  25 &
  \cellcolor[HTML]{969696}5 &
  25 &
  \cellcolor[HTML]{BBBBBB}21 &
  \cellcolor[HTML]{FFFFFF}25 &
  \cellcolor[HTML]{969696}13 \\ \hline
Dataset &
  \multicolumn{6}{c|}{Bank: Age} &
  \multicolumn{6}{c|}{German: Sex} &
  \multicolumn{6}{c|}{Health: Age} &
  \multicolumn{6}{c|}{MEPS: Race} \\ \hline
Method &
  XGB &
  EGR &
  FAXAI &
  FairMask &
  MAAT &
  \textbf{FairMatch} &
  XGB &
  EGR &
  FAXAI &
  FairMask &
  MAAT &
  \textbf{FairMatch} &
  XGB &
  EGR &
  FAXAI &
  FairMask &
  MAAT &
  \textbf{FairMatch} &
  XGB &
  EGR &
  FAXAI &
  FairMask &
  MAAT &
  \textbf{FairMatch} \\ \hline
Accuracy &
  \cellcolor[HTML]{969696}91 &
  \cellcolor[HTML]{969696}91 &
  \cellcolor[HTML]{969696}91 &
  \cellcolor[HTML]{969696}91 &
  \cellcolor[HTML]{969696}90 &
  \cellcolor[HTML]{969696}91 &
  \cellcolor[HTML]{969696}75 &
  \cellcolor[HTML]{969696}75 &
  \cellcolor[HTML]{969696}75 &
  \cellcolor[HTML]{969696}75 &
  \cellcolor[HTML]{969696}77 &
  \cellcolor[HTML]{969696}75 &
  \cellcolor[HTML]{969696}78 &
  \cellcolor[HTML]{969696}78 &
  \cellcolor[HTML]{969696}78 &
  \cellcolor[HTML]{969696}78 &
  \cellcolor[HTML]{969696}79 &
  \cellcolor[HTML]{969696}79 &
  \cellcolor[HTML]{969696}87 &
  \cellcolor[HTML]{969696}86 &
  \cellcolor[HTML]{969696}87 &
  \cellcolor[HTML]{969696}86 &
  \cellcolor[HTML]{969696}86 &
  \cellcolor[HTML]{969696}85 \\
Precision &
  \cellcolor[HTML]{BBBBBB}67 &
  \cellcolor[HTML]{BBBBBB}67 &
  \cellcolor[HTML]{BBBBBB}67 &
  \cellcolor[HTML]{BBBBBB}67 &
  \cellcolor[HTML]{969696}76 &
  \cellcolor[HTML]{BBBBBB}66 &
  \cellcolor[HTML]{969696}78 &
  \cellcolor[HTML]{969696}79 &
  \cellcolor[HTML]{969696}79 &
  \cellcolor[HTML]{969696}78 &
  \cellcolor[HTML]{969696}67 &
  \cellcolor[HTML]{969696}79 &
  \cellcolor[HTML]{969696}78 &
  \cellcolor[HTML]{969696}79 &
  \cellcolor[HTML]{969696}78 &
  \cellcolor[HTML]{969696}79 &
  \cellcolor[HTML]{969696}78 &
  \cellcolor[HTML]{969696}78 &
  \cellcolor[HTML]{969696}65 &
  \cellcolor[HTML]{969696}68 &
  \cellcolor[HTML]{969696}67 &
  \cellcolor[HTML]{969696}66 &
  \cellcolor[HTML]{969696}67 &
  \cellcolor[HTML]{BBBBBB}53 \\
Recall &
  \cellcolor[HTML]{BBBBBB}55 &
  \cellcolor[HTML]{BBBBBB}53 &
  \cellcolor[HTML]{BBBBBB}55 &
  \cellcolor[HTML]{BBBBBB}55 &
  \cellcolor[HTML]{969696}80 &
  \cellcolor[HTML]{BBBBBB}55 &
  \cellcolor[HTML]{969696}88 &
  \cellcolor[HTML]{969696}88 &
  \cellcolor[HTML]{969696}88 &
  \cellcolor[HTML]{969696}88 &
  \cellcolor[HTML]{969696}72 &
  \cellcolor[HTML]{969696}87 &
  \cellcolor[HTML]{969696}76 &
  \cellcolor[HTML]{969696}75 &
  \cellcolor[HTML]{969696}75 &
  \cellcolor[HTML]{969696}77 &
  \cellcolor[HTML]{969696}79 &
  \cellcolor[HTML]{969696}76 &
  \cellcolor[HTML]{BBBBBB}42 &
  \cellcolor[HTML]{FFFFFF}36 &
  \cellcolor[HTML]{BBBBBB}41 &
  \cellcolor[HTML]{BBBBBB}42 &
  \cellcolor[HTML]{969696}78 &
  \cellcolor[HTML]{BBBBBB}50 \\
F1 &
  \cellcolor[HTML]{BBBBBB}60 &
  \cellcolor[HTML]{BBBBBB}59 &
  \cellcolor[HTML]{BBBBBB}60 &
  \cellcolor[HTML]{BBBBBB}60 &
  \cellcolor[HTML]{969696}78 &
  \cellcolor[HTML]{BBBBBB}60 &
  \cellcolor[HTML]{969696}83 &
  \cellcolor[HTML]{969696}83 &
  \cellcolor[HTML]{969696}83 &
  \cellcolor[HTML]{969696}83 &
  \cellcolor[HTML]{969696}69 &
  \cellcolor[HTML]{969696}82 &
  \cellcolor[HTML]{969696}76 &
  \cellcolor[HTML]{969696}76 &
  \cellcolor[HTML]{969696}76 &
  \cellcolor[HTML]{969696}78 &
  \cellcolor[HTML]{969696}78 &
  \cellcolor[HTML]{969696}77 &
  \cellcolor[HTML]{BBBBBB}51 &
  \cellcolor[HTML]{FFFFFF}47 &
  \cellcolor[HTML]{BBBBBB}51 &
  \cellcolor[HTML]{BBBBBB}52 &
  \cellcolor[HTML]{969696}70 &
  \cellcolor[HTML]{BBBBBB}51 \\
AOD &
  \cellcolor[HTML]{BBBBBB}4 &
  \cellcolor[HTML]{969696}3 &
  \cellcolor[HTML]{969696}3 &
  \cellcolor[HTML]{969696}3 &
  \cellcolor[HTML]{969696}3 &
  \cellcolor[HTML]{969696}3 &
  \cellcolor[HTML]{969696}3 &
  \cellcolor[HTML]{969696}4 &
  \cellcolor[HTML]{969696}4 &
  \cellcolor[HTML]{969696}3 &
  \cellcolor[HTML]{969696}6 &
  \cellcolor[HTML]{969696}2 &
  \cellcolor[HTML]{969696}6 &
  11 &
  11 &
  \cellcolor[HTML]{969696}6 &
  \cellcolor[HTML]{FFFFFF}13 &
  \cellcolor[HTML]{BBBBBB}9 &
  \cellcolor[HTML]{BBBBBB}5 &
  \cellcolor[HTML]{969696}3 &
  \cellcolor[HTML]{BBBBBB}6 &
  \cellcolor[HTML]{BBBBBB}5 &
  \cellcolor[HTML]{BBBBBB}5 &
  \cellcolor[HTML]{969696}4 \\
EOD &
  10 &
  \cellcolor[HTML]{BBBBBB}7 &
  \cellcolor[HTML]{969696}6 &
  \cellcolor[HTML]{BBBBBB}8 &
  \cellcolor[HTML]{969696}5 &
  \cellcolor[HTML]{969696}6 &
  \cellcolor[HTML]{969696}3 &
  \cellcolor[HTML]{BBBBBB}5 &
  \cellcolor[HTML]{969696}4 &
  \cellcolor[HTML]{969696}3 &
  \cellcolor[HTML]{969696}4 &
  \cellcolor[HTML]{969696}4 &
  16 &
  \cellcolor[HTML]{BBBBBB}11 &
  \cellcolor[HTML]{969696}9 &
  \cellcolor[HTML]{969696}9 &
  \cellcolor[HTML]{BBBBBB}12 &
  \cellcolor[HTML]{BBBBBB}14 &
  \cellcolor[HTML]{BBBBBB}15 &
  \cellcolor[HTML]{969696}6 &
  \cellcolor[HTML]{BBBBBB}15 &
  \cellcolor[HTML]{969696}8 &
  \cellcolor[HTML]{969696}8 &
  \cellcolor[HTML]{969696}5 \\
SPD &
  11 &
  \cellcolor[HTML]{969696}8 &
  \cellcolor[HTML]{BBBBBB}9 &
  \cellcolor[HTML]{BBBBBB}10 &
  \cellcolor[HTML]{969696}8 &
  \cellcolor[HTML]{969696}8 &
  \cellcolor[HTML]{969696}4 &
  \cellcolor[HTML]{BBBBBB}6 &
  \cellcolor[HTML]{969696}4 &
  \cellcolor[HTML]{969696}2 &
  \cellcolor[HTML]{969696}5 &
  \cellcolor[HTML]{969696}3 &
  \cellcolor[HTML]{BBBBBB}30 &
  \cellcolor[HTML]{969696}26 &
  \cellcolor[HTML]{969696}26 &
  \cellcolor[HTML]{969696}25 &
  \cellcolor[HTML]{969696}23 &
  \cellcolor[HTML]{969696}24 &
  12 &
  \cellcolor[HTML]{969696}5 &
  \cellcolor[HTML]{BBBBBB}10 &
  \cellcolor[HTML]{BBBBBB}10 &
  \cellcolor[HTML]{BBBBBB}8 &
  \cellcolor[HTML]{969696}7 \\
DI &
  105 &
  \cellcolor[HTML]{969696}78 &
  \cellcolor[HTML]{BBBBBB}93 &
  \cellcolor[HTML]{BBBBBB}99 &
  \cellcolor[HTML]{FFFFFF}154 &
  \cellcolor[HTML]{969696}80 &
  \cellcolor[HTML]{BBBBBB}5 &
  7 &
  \cellcolor[HTML]{BBBBBB}5 &
  \cellcolor[HTML]{969696}3 &
  \cellcolor[HTML]{969696}5 &
  \cellcolor[HTML]{969696}4 &
  50 &
  \cellcolor[HTML]{BBBBBB}46 &
  \cellcolor[HTML]{BBBBBB}44 &
  \cellcolor[HTML]{BBBBBB}45 &
  \cellcolor[HTML]{969696}39 &
  \cellcolor[HTML]{969696}41 &
  63 &
  \cellcolor[HTML]{969696}39 &
  60 &
  \cellcolor[HTML]{BBBBBB}56 &
  \cellcolor[HTML]{BBBBBB}52 &
  \cellcolor[HTML]{969696}38 \\ \hline
\end{tabular}%
}
\end{table*}

\begin{blockquote}
\textbf{RQ4}: Can FairMatch obtain superior trade-offs between performance and fairness?
\end{blockquote}
Since PSM can be used to match comparable data samples, we decided to use this procedure to develop a post-processing method to mitigate bias. 
Table \ref{rq3-1} and Table \ref{rq3-2} compare our methods with 4 other baselines that are prior state-of-the-art works. 
As previously introduced in \S\ref{background}, exponentiated gradient reduction (EGR) is an in-processing technique that applies to most classification algorithms supported in scikit-learn. Fax-AI and FairMask are both post-processing methods that mitigate algorithmic discrimination by limiting the direct/indirect influence of protected attributes on the decision model. MAAT is an ensemble learning framework that aims to achieve better performance-fairness trade-off by training and aggregating different models, respectively, oriented by predictive performance and fairness constraints. 
As reported in both Table \ref{rq3-1} and Table \ref{rq3-2}, it is obvious that our method can achieve similar or superior fairness-performance trade-offs on both base models (logistic regression and gradient-based boosting tree). It is also noteworthy that MAAT can achieve much better performance scores than other mitigation methods in certain datasets (Adult, Bank, and MEPS).

Overall, our answer to RQ4 is {\bf In the majority of cases, the new bias mitigation method based on PSM performs better or is similar to other state-of-the-art algorithms in terms of both fairness and performance.}

\section{Threats to Validity}\label{threats}
\label{threat}

\noindent
\textbf{Sampling Bias} Most of the prior works~\cite{,chakraborty2019software,NIPS2017_6988,Galhotra_2017,zhang2018mitigating,Kamiran:2018:ERO:3165328.3165686} used one or two datasets where we used six well-known datasets in our experiments. There are also other datasets being collected and released in the community of fairness study. In the future, we will extend our experimentation on more datasets and learners. \\
\textbf{Evaluation Bias} - We used the four fairness metrics in this study, compared to previous works that \cite{Chakraborty_2020,10.1007/978-3-642-33486-3_3,hardt2016equality} used fewer metrics. However. IBM AIF360~\cite{bellamy2018ai} contains more than 50 metrics and counting. Also, this paper does not include metrics that measure individual fairness. More evaluation criteria will be examined in future work.\\
\textbf{Conclusion Validity} - Our approach is designed for batch-sampled fairness testing. That is, we assume in the offline testing stage of software, data samples will always come in representative batches, from which point PSM can be conducted. Therefore, this might not apply to online testing or real-time bias mitigation, especially when incoming data is sparse and sporadic. \\
\textbf{Instrumental Validity} - PSM relies on the estimated propensity scores to determine comparable samples. Therefore, the "smoothness" of the base model significantly influences the sampling result as it is used to identify neighbors. For example, if the base model cannot provide predicted probability scores with enough granularity, this will result in too many individuals being classified as neighbors of each other. In our study, we experimented with several candidate models and selected models capable of providing smooth propensity scores.\\
\textbf{External Validity} - Our work is limited to binary classification and tabular data, which are very common in AI software. However, all the methods used in this paper can easily be extended for multi-class classification and regression problems. In the future, we will try to extend our work to other domains of SE and ML.

\section{Conclusion}\label{conclusion}
Fairness testing plays an increasingly crucial role in ML software development nowadays. Various fairness metrics are defined based on different intuitions of ensuring distributive fairness. However, the manner in which these fairness metrics are used in the model testing phase is under-explored, especially when the training data is speculated/detected to contain intentional discrimination. This paper argues that traditional fairness testing schema may not always reflect a model's actual fairness/bias level. Instead, the fairness scores highly depend on whether the "right" test set is used. To resolve this open issue, we proposed a propensity-based fairness testing schema. Following our new testing pipeline, one can receive a more comprehensive analysis of how the fairness scores may vary as the distribution of protected attributes changes among the testing samples.

% Our results offer some bad news and some good news.
% The bad news is that many prior works on fairness testing had an undetected threat to validity. That prior work, which ignored propensity issues, needs to be revisited since that work may have used the wrong test data. 

Beyond fairness testing, this paper has proposed a bias mitigation method, FairMatch, that leverages propensity score matching to locate samples with a higher probability of being discriminated by the decision model. In experiments, we found that the proposed approach can achieve on-par or superior fairness-performance trade-offs compared to benchmark methods.
In summary, we conclude that:
\begin{itemize}
    \item We can report that fairness testing is very sensitive to the choice of sampling strategies, which may lead to a wide spectrum of trade-offs between performance and fairness.
    \item We can provide a novel fairness testing mechanism, which retires a potential threat to validity.
    \item We can use propensity score matching (PSM) to locate the subset of samples with a greater risk of algorithmic discrimination.
    \item We can recommend using FairMatch as a post-processing tool to efficiently mitigate bias without requesting to update the original model.
\end{itemize}

\section*{Acknowledgements}
This work was partially funded by 
a research grant from the Laboratory for Analytical Sciences, North Carolina State University.
% blinded for review.

% \bibliographystyle{reference}
% \bibliography{reference.bib}
% \bibliographystyle{IEEEtran}
\bibliographystyle{elsarticle-num}

\bibliography{main}

% \clearpage
% \setcounter{page}{1}
% \pagenumbering{roman}
% \normalsize
% \twocolumn
% \newpage
% \twocolumn

\end{document}